\documentclass{article}

\usepackage[preprint]{neurips_2022}

\usepackage{listings}

\usepackage[utf8]{inputenc} 
\usepackage[T1]{fontenc}    
\usepackage{hyperref}       
\usepackage{url}            
\usepackage{booktabs}       
\usepackage{amsfonts}       
\usepackage{nicefrac}       
\usepackage{microtype}      
\usepackage{xcolor}         

\usepackage{natbib}
\usepackage{adjustbox}

\usepackage{amsmath}
\usepackage{spverbatim}

\usepackage{subcaption}

\newcommand{\pretrained}[0]{{\texttt{Pre-trained\_T5}}}
\newcommand{\randominit}[0]{{\texttt{Random\_Init}}}
\newcommand{\wiki}[0]{{\texttt{Wiki\_10K}}}

\newcommand{\set}[0]{{\texttt{Set}}}
\newcommand{\identity}[0]{{\texttt{Identity}}}

\newcommand{\lime}[0]{{\texttt{LIME}}}
\newcommand{\nonsensesummary}[0]{{\texttt{Nonsense\_Summary}}}

\newcommand{\nonsensesummaryshort}[0]{{\texttt{Nons\_Summary}}}

\newcommand{\dyck}[0]{{\texttt{Dyck}}}

\newcommand{\perparammeansd}[0]{{\texttt{Per\_Param\_Mean\_SD}}}
\newcommand{\perparamscale}[0]{{\texttt{Per\_Param\_Scale}}}
\newcommand{\wholemodelscale}[0]{{\texttt{Whole\_Model\_Scale}}}
\newcommand{\subsetperparamscale}[0]{{\texttt{Subset\_Per\_Param\_Scale}}}
\newcommand{\preattnlnsubsetperparamscale}[0]{{\texttt{Pre-attn\_LN\_Per\_Param\_Scale}}}
\newcommand{\fullinit}[0]{{\texttt{Full\_Init}}}

\newcommand{\acrosslayers}[0]{{\texttt{Across\_Layers\_Scale}}}
\newcommand{\perlayer}[0]{{\texttt{Per\_Layer\_Scale}}}

\newcommand{\customfootnotetext}[2]{{
  \renewcommand{\thefootnote}{#1}
  \footnotetext[0]{#2}}}
\definecolor{citecolor}{rgb}{.259,.659,1}
\definecolor{mydarkblue}{rgb}{0,0.08,0.45}
\definecolor{urlcolor}{rgb}{0,.145,.698}
\definecolor{linkcolor}{rgb}{.71,0.21,0.01}
\hypersetup{
    colorlinks=true,
    breaklinks=true,  
    bookmarksnumbered=true,
    linkcolor=linkcolor,
    citecolor=citecolor,
    filecolor=mydarkblue,
    urlcolor=urlcolor,
    pdftitle={Insights into Pre-training via Simpler Synthetic Tasks},
    pdfpagemode=FullScreen,
    pdfview=FitH
    }

\usepackage{tcolorbox}
\usepackage{listings}

\usepackage[style=1]{mdframed}

\definecolor{codegreen}{rgb}{0,0.6,0}
\definecolor{codegray}{rgb}{0.5,0.5,0.5}
\definecolor{codepurple}{rgb}{0.58,0,0.82}
\definecolor{backcolour}{rgb}{0.95,0.95,0.92}

\lstdefinestyle{mystyle}{
    commentstyle=\color{codegreen},
    keywordstyle=\color{magenta},
    stringstyle=\color{codepurple},
    basicstyle=\ttfamily\scriptsize,
    breakatwhitespace=false,         
    breaklines=false,                 
    captionpos=b,                    
    keepspaces=true,                 
    showspaces=false,                
    showstringspaces=false,
    showtabs=false,                  
    tabsize=2,
    basewidth=0.5em
}

\mdfdefinestyle{MyFrame}{%
    linecolor=codegray,
    outerlinewidth=1pt,
    innertopmargin=4pt,
    innerbottommargin=4pt,
    innerrightmargin=4pt,
    innerleftmargin=4pt,
        leftmargin = 4pt,
        rightmargin = 4pt
        }

\lstnewenvironment{example2}[1]
{
    \mdframed[style=MyFrame,nobreak=true]
    \vspace{0.05in}
    \textbf{\footnotesize Input:}
}
{
    \textbf{\footnotesize Output:}\vspace{0.05in}\newline\ttfamily\scriptsize \texttt{#1}
    \endmdframed
}

\lstnewenvironment{examplemiddle}[1]
{
    \mdframed[style=MyFrame,nobreak=true]
    \vspace{0.05in}
    \textbf{\footnotesize Input:}
}
{
    \vspace{0.67in}\textbf{\footnotesize Output:}\vspace{0.05in}\newline\ttfamily\scriptsize \texttt{#1}
    \endmdframed
}

\title{Insights into Pre-training via Simpler Synthetic Tasks}

\author{%
  Yuhuai Wu$^{12}$\thanks{Equal Contribution.} \\
  \texttt{yuhuai@cs.stanford.edu} \\
  \And
  Felix Li$^{3*}$ \\
  \texttt{fzli@berkeley.edu} \\
  \And
  Percy Liang$^{1}$ \\
  \texttt{pliang@cs.stanford.edu} 
  \AND \normalfont{$^1$Stanford University} \\ $^2$Google Research  \\$^3$UC Berkeley}

\begin{document}
\maketitle
\begin{abstract}
Pre-training produces representations that are effective for a wide range of downstream tasks, but it is still unclear what properties of pre-training are necessary for effective gains. Notably, recent work shows that even pre-training on synthetic tasks can achieve significant gains in downstream tasks. In this work, we perform three experiments that iteratively simplify pre-training and show that the simplifications still retain much of its gains. First, building on prior work, we perform a systematic evaluation of three existing synthetic pre-training methods on six downstream tasks. We find the best synthetic pre-training method, \lime{}, attains an average of $67\%$ of the benefits of natural pre-training. Second, to our surprise, we find that pre-training on a simple and generic synthetic task defined by the \set{} function achieves $65\%$ of the benefits, almost matching \lime{}. Third, we find that $39\%$ of the benefits can be attained by using merely the parameter statistics of synthetic pre-training. We release the source code at \url{https://github.com/felixzli/synthetic_pretraining}.
\end{abstract}

\section{Introduction}

Pre-training on a large amount of data from sources such as text from the web---\emph{natural pre-training}---is effective for a wide range of downstream tasks~\citep{Devlin2019bert, brown2020gpt3,chen2021codex, codexglue}.
More surprisingly, recent works~\citep{lime, nonsense_summary, ri,fractal} show that pre-training on data that is fully synthetically generated---\emph{synthetic pre-training}---can also provide substantial gains, partially closing the gap between training from a randomly initialized model and a naturally pre-trained model. 

\begin{figure}[t]
\centering 
\centerline{\includegraphics[width=0.8\columnwidth]{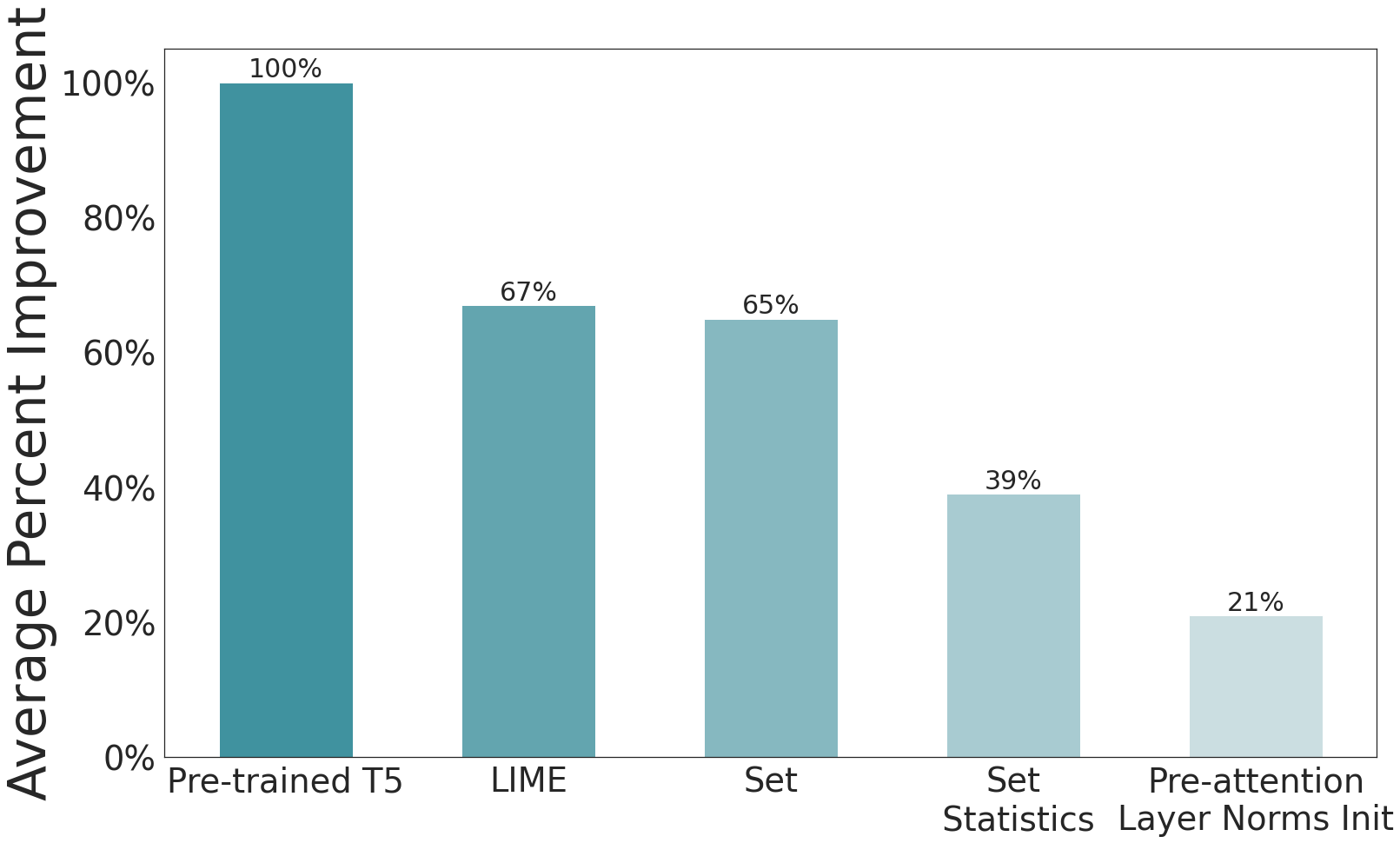}}
\vspace{+0.1in}
\caption{
We compare downstream task performance between natural language pre-training (\pretrained{}), synthetic pre-training (\lime{}), simpler synthetic pre-training (\set{}), initializing from the statistics of Set pre-training (\set{} Statistics), and initializing just the pre-attention layer norms to a lower value (Pre-attention Layer Norm Init). The average percent improvement is computed by taking the average over our six downstream tasks of $\frac{\text{Performance(Initialization)} - \text{Performance(\randominit{})}}{\text{Performance(\pretrained{})}-\text{Performance(\randominit{}})}$.
}
\vspace{-0.1in}
\label{section1:figure}
\end{figure}

What properties of pre-training are necessary for effective gains? In this paper, we conduct a careful empirical study to help understand this question. We perform three experiments that iteratively simplify natural pre-training, while still retaining much of the benefits of natural pre-training. See Figure~\ref{section1:figure} for an overview of the results obtained in our three experiments.

First, we perform a systematic evaluation of three previously proposed synthetic pre-training methods over six downstream tasks. Prior works only evaluated each synthetic pre-training method on a single downstream domain: \citet{lime} on mathematical reasoning benchmarks, \citet{chiang} on GLUE, \citet{nonsense_summary} on summarization, and \citet{ri} on language modeling and dependency parsing. We instead show how general these methods are as well as how different synthetic pre-training methods compare against each other. We find that the best performing synthetic task, \lime{}~\citep{lime}, closes a significant fraction of the gap between random initialization and natural pre-training for summarization ($85\%$), semantic parsing ($83\%$), reading comprehension ($55\%$), code translation ($43\%$), and retrosynthesis ($49\%$).

Second, we discover a surprisingly simple and generic synthetic task that captured nearly the same effectiveness provided by the existing best synthetic task \lime{}, closing an average of $65\%$ of the gap to natural pre-training (compared to $67\%$ for \lime{}). This simple task, which we call \set{}, consists of outputting the input sequence without duplicate tokens in its original order, for example:
\begin{align*}
    \text{Input: a a c b a b} \Rightarrow \text{Output: a c b}.
\end{align*}

Third, to simplify the initialization even more, we ask whether only the statistics of the pre-trained weights can explain the effectiveness of synthetic pre-training.
We find that initializing parameters with statistics of a synthetic pre-trained model provides large benefits for some downstream tasks, closing $39\%$ of the gap to natural pre-training. Simplifying further allows us to see that by initializing all pre-attention layer norm parameters to a lower value (just one number!), we close more than $40\%$ of the gap to natural pre-training performance for summarization and semantic parsing, and an average of $21\%$ over all six tasks.
\section{Experimental Setup}
\label{section2:setup}



\paragraph{Architecture}
We trained a T5-Small model, which is a 60 million parameter encoder-decoder transformer model with 6 encoder layers, 6 decoder layers, 8 heads, head dimension 64, and MLP dimension 2048 \citep{t5}. 

\paragraph{Training Details}
For synthetic pre-training, we use the same hyperparameters that the off-the-shelf language pre-trained T5-small was trained with: AdaFactor optimizer, batch size 128, sequence length 512, and inverse square root learning rate $1/\sqrt{\max(n, 10000)}$ where $n$ is the current training step. We evaluate token validation accuracy every 5000 training steps. For all synthetic tasks besides \dyck{}, we fine-tuned with the first checkpoint that the model reaches above 99\% token validation accuracy. For \dyck{}, the model's validation accuracy plateaued without reaching above 99\%, so we chose to fine-tune using the checkpoint with max validation accuracy of 77.7\%. We provide further pre-training details in Appendix~\ref{appendix:pretrain_details} and fine-tuning details in Appendix~\ref{appendix:finetune_details}.

\paragraph{Downstream Tasks}
We fine-tuned synthetically pre-trained models on a diverse suite of downstream tasks: 1) Java to C\# code translation (10K training examples) \citep{codexglue}; 2) two semantic parsing benchmarks, MTOP (17K training examples) \citep{mtop} and WebQSP (2.7K training examples) \citep{webqsp}, consisting of converting a natural language query to a logical form; 3) USPTO-50K retrosynthesis (40K training examples) \citep{uspto-50K}, a task that consists of predicting possible reactants when given a product as input; 4) the reading comprehension benchmark SQuAD 1.1 (87K training examples) \citep{squad}; and 5) the summarization benchmark CNNDM-10K$^{3}$ which is 10K training examples from the CNNDM  \citep{nonsense_summary}. See full descriptions and input-output examples for every downstream task in Appendix~\ref{appendix:downstream_tasks}.

\paragraph{Baselines}
We compared against randomly initialized (\randominit{}) and off-the-shelf natural language pre-trained T5-Small (\pretrained{}). \pretrained{} was trained for 524K steps on the C4 dataset with the span-corruption objective \citep{t5}. We also compared against T5-Small trained on Wikipedia \citep{wikipedia} for 10K steps (\wiki{}) with the same pre-training objective as \pretrained{}. 

\paragraph{Source code} We release the source code to reproduce the experiments at \url{https://github.com/felixzli/synthetic_pretraining}.
\section{Benchmarking Existing Synthetic Pre-training}
\label{section3}

\begin{table}[]
\caption{Evaluation of three previously proposed synthetic pre-training tasks and two simpler synthetic pre-training tasks. For tasks where we report both Exact Match (EM) and F1, the Average column is computed using EM.
}
  \label{section3:table_main}
\begin{adjustbox}{width=1.0\linewidth}
\begin{tabular}{@{}ccccccccc@{}}
                & CNNDM-10K & MTOP      & WebQSP    & SQuAD     & Code Trans. & \multicolumn{1}{c|}{Retrosyn.} & Average \\
                & ROUGE1    & EM/F1     & EM/F1     & EM/F1     & EM          & \multicolumn{1}{c|}{EM}        &         \\ \midrule
\multicolumn{8}{c}{Baselines}                                                                                            \\ \midrule
\pretrained{}  & \textbf{35.8}$^{1}$ & \textbf{81.0/95.2} & \textbf{83.1/91.9} & \textbf{77.5/86.0} & \textbf{61.6} & \multicolumn{1}{c|}{\textbf{43.1}} & \textbf{63.7} & \textbf{100\%} \\
Wiki 10K        & 34.0      & 71.6/94.2 & 79.6/90.6 & 67.1/76.9 & 60.2        & \multicolumn{1}{c|}{41.1}      & 58.9  & 77\% \\
\randominit{} & 18.9 & 38.6/83.0 & 28.2/72.9 & 17.2/25.2 & 57.0 & \multicolumn{1}{c|}{39.2} & 33.2 & \textbf{0\%} \\ \midrule
\multicolumn{8}{c}{Section~\ref{section3}: Benchmarking Existing Synthetic Pre-training}                                                         \\ \midrule
\lime{}        & \textbf{33.2} & \textbf{73.7/94.0} & \textbf{75.2/88.7} & \textbf{50.4/62.1} & \textbf{59.0} & \multicolumn{1}{c|}{\textbf{41.1}} & \textbf{55.4} & \textbf{67\%}\\
\dyck{}  & 27.1      & 65.9/91.9 & 58.5/83.5 & 50.3/62.4 & 58.8        & \multicolumn{1}{c|}{40.4}      & 50.2   & 49\% \\
\nonsensesummaryshort{}$^{2}$ & 32.0      & 68.0/92.7 & 65.2/85.2 & 48.4/60.1 & 57.3        & \multicolumn{1}{c|}{39.6}      & 51.8  & 47\%  \\ \midrule
\multicolumn{8}{c}{Section~\ref{section4}: Simpler Synthetic Pre-Training}                                                                     \\ \midrule
\set{}          & \textbf{32.8} & \textbf{71.7/93.7} & \textbf{72.2/88.2} & \textbf{48.3/60.0} & \textbf{59.4} & \multicolumn{1}{c|}{\textbf{40.9}} & \textbf{54.2}& \textbf{65\%} \\
\identity{}           & 30.2      & 68.6/93.0 & 69.8/86.8 & 26.1/35.4 & 57.8        & \multicolumn{1}{c|}{40.5}      & 48.8 & 46\%    \\ \bottomrule
\end{tabular}
\end{adjustbox}
\end{table}

We benchmark three previously proposed synthetic pre-training tasks across six downstream tasks. We compute how much of the benefits using natural pre-training are gained by each synthetic pre-training method by taking the average over six downstream tasks of $\frac{\text{Performance(Initialization)} - \text{Performance(\randominit{})}}{\text{Performance(\pretrained{})}-\text{Performance(\randominit{}})}$. We show these numbers in the right-most column of Table~\ref{section3:table_main}. We find that \lime{} achieves the best performance, gaining 67\% of the benefits using natural pre-training.

\subsection{Synthetic Tasks}
\label{section3:synthetic_tasks}
\begin{figure}[t]
\centering 
\centerline{\includegraphics[width=1.0\columnwidth]{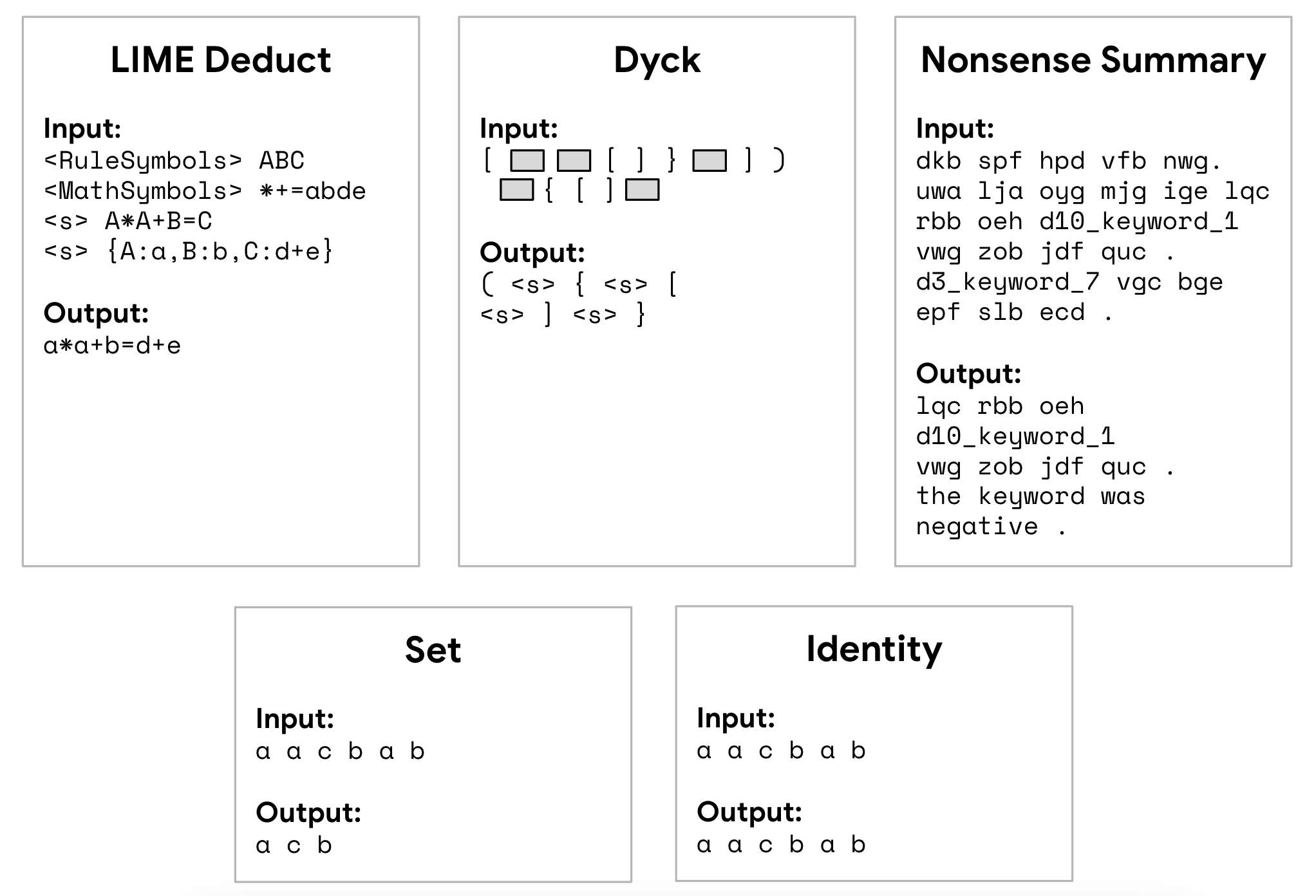}}
\caption{
We show examples of Section~\ref{section3} previously proposed synthetic tasks in the top row and Section~\ref{section4} simpler synthetic tasks in the bottom row. \textbf{Top Left:} \lime{} deduct task, whose input consists of a "rule string" and "substitution dictionary" and output is a "result string." The "rule symbols" and "math symbols" are written as actual letters and math operations for reader clarity. In real generated data, each example's rule and math symbols are randomly sampled tokens. \textbf{Top Middle:} T5 style masked language modelling with \dyck{} language. In the input, the grey boxes indicate masked out spans. The model is trained to predict the masked spans, each separated by a special token \texttt{<s>}. In this example, the "noise\_density" is 0.33 (so 5 out of 15 input brackets are corrupted) and the "span\_corruption\_length" is 1 (so each masked span is of length 1). In actual generated data, the "span\_corruption\_length" is 3 and the "noise\_density" is 0.15, which are the same parameter values used for natural language \pretrained{}. \textbf{Top Right:} \nonsensesummary{} task. In this example, two operations to create the summary are "copy sentence containing a keyword" and "identify keyword sentiment." \textbf{Bottom Left:} \set{} task. \textbf{Bottom Right:} \identity{} task.}
\vspace{-0.1in}
\label{section3:figure_examples}
\end{figure}

For each synthetic task, we generated one million examples. To generate synthetic task data, we sampled tokens a vocabulary of size 32K, matching the vocabulary size of \pretrained{}. Details about data generation can be found in the original papers. We provide an example of each task in Figure~\ref{section3:figure_examples}.

\customfootnotetext{$1$}{This number was obtained using pre-trained \texttt{T5\_1.1} instead of \texttt{T5\_1.0} which we used for all other tasks. 
Using pre-trained \texttt{T5\_1.0} would be an unfair comparison because it has already been trained on all 290K CNNDM task training data. Full differences between \texttt{T5\_1.1} and \texttt{T5\_1.0} are described in Appendix Section~\ref{appendix:T5v1.1}.}

\customfootnotetext{$2$}{We obtained different results than \citet{nonsense_summary} on CNNDM-10K because of our choice of optimizer. We were able to match their results when we used Adam like they did in their work instead of Adafactor. We chose to use Adafactor instead of Adam because training took about three times less steps.}
\customfootnotetext{$3$}{We followed \citet{nonsense_summary} to evaluate on CNNDM-10K instead of the full CNNDM. They used CNNDM-10K because less training examples makes the impact of pre-training more visible.}

\paragraph{LIME}
\lime{} is a set of three tasks---Deduct, Induct, and Abduct---inspired by Charles Peirce's three reasoning primitives \citep{lime}. Each task consists of three elements: a rule string, a dictionary that represents substitutions, and a result string that is the result of applying those substitutions to the rule string. The three tasks are then constructed by using two of the three elements as inputs to predict the remaining element.

\paragraph{Dyck Artificial Language} 
We consider a Dyck language with multiple bracket types (taken from \citet{ri}), which we refer as \dyck{}.
We turn generated text into a sequence-to-sequence task by using the T5 span-corruption pre-training objective~\citep{t5}. 
The motivation behind the design of \dyck{} is how ``sentences of natural language often have dependency relations where the existence of a certain word can be predictive of another word’’ \citep{ri}.

\paragraph{Nonsense Summary}
\citet{nonsense_summary} developed the nonsense summarization task, which we refer to as \nonsensesummary{}, by conducting a qualitative analysis of summarization data and identifying 21 elementary operations used to create summaries such as ``CopyFirstSentence’’ and ``CopyQuoted.’’ Each synthetic task example consists of an input that is a nonsense document made of randomly sampled tokens and an output that is the corresponding nonsense summary consisting of applying a random subset of the 21 operations to the document. 

\subsection{Results}
\label{section3:results}
Table~\ref{section3:table_main} shows results of evaluating the synthetic tasks across six downstream tasks, and Appendix Table~ \ref{appendix:table_dataset_size_ablation} shows results of ablating the dataset size for retrosynthesis, CNNDM, and SQuAD.

\paragraph{Synthetic pre-training provides large gains over random initialization.}
We observed large gains from all three synthetic tasks we evaluated. Out of these, \lime{} provided the most benefits across all six downstream tasks. We computed how much of the natural pre-training benefit is captured by \lime{} as $\frac{\text{Performance(\lime{})} - \text{Performance(\randominit{})}}{\text{Performance(\pretrained{})}-\text{Performance(\randominit{}})}$. \lime{} closed $84.6\%$, $82.8\%$, $85.6\%$, $55.1\%$, $43.5\%$,	and $48.7\%$ of the gap for CNNDM-10K, MTOP, WebQSP, SQuAD, Code Translation, and Retrosynthesis respectively.

\paragraph{Task or domain prior is not required for pre-training benefits.}
For example, \lime{} benefits tasks including summarization (CNNDM-10K) and reading comprehension (SQuAD) despite \lime{} pre-training data involving no natural language like \pretrained{}, representing no structure of natural language like \dyck{}, and not being explicitly designed to reflect the operations used in summarization like \nonsensesummary{}.

\paragraph{Varying degrees of benefits.} 
Comparing \nonsensesummary{} and \dyck{}, we see that the two synthetic tasks benefit MTOP, WebQSP, and SQuAD similarly, yet their benefits for the three other tasks are very different.
The difference between \nonsensesummary{} and \dyck{} as well as the superiority of \lime{} are examples of how different synthetic pre-training can have significantly different effectiveness.

\paragraph{Synthetic pre-training is not more efficient.}
One of the advantages argued for synthetic pre-training is its computational efficiency ~\citep{lime,nonsense_summary}. To explore this potential advantage, we pre-trained our model using the T5 span-corruption pre-training objective on Wikipedia~\citep{wikipedia} for 10K steps, and the result is shown in Table~\ref{section3:table_main} labelled \wiki{}. Wikipedia pre-training for 10K steps provided more significant gains than \lime{} pre-training for 30K steps and Set pre-training for 10K steps. Based on this result, we see no evidence that the evaluated synthetic pre-training tasks provides computational efficiency benefits over natural language pre-training, unlike what \citet{lime} suggested.

\paragraph{Natural language pre-training benefits non-natural language tasks.}
Natural language pre-training, including \wiki{}, outperforms synthetic pre-training on the non-natural language tasks of Code Translation and Retrosynthesis. 
\section{Simpler Synthetic Pre-training}
\label{section4}

\begin{figure}[t]
\centering 
\centerline{\includegraphics[width=1.0\columnwidth]{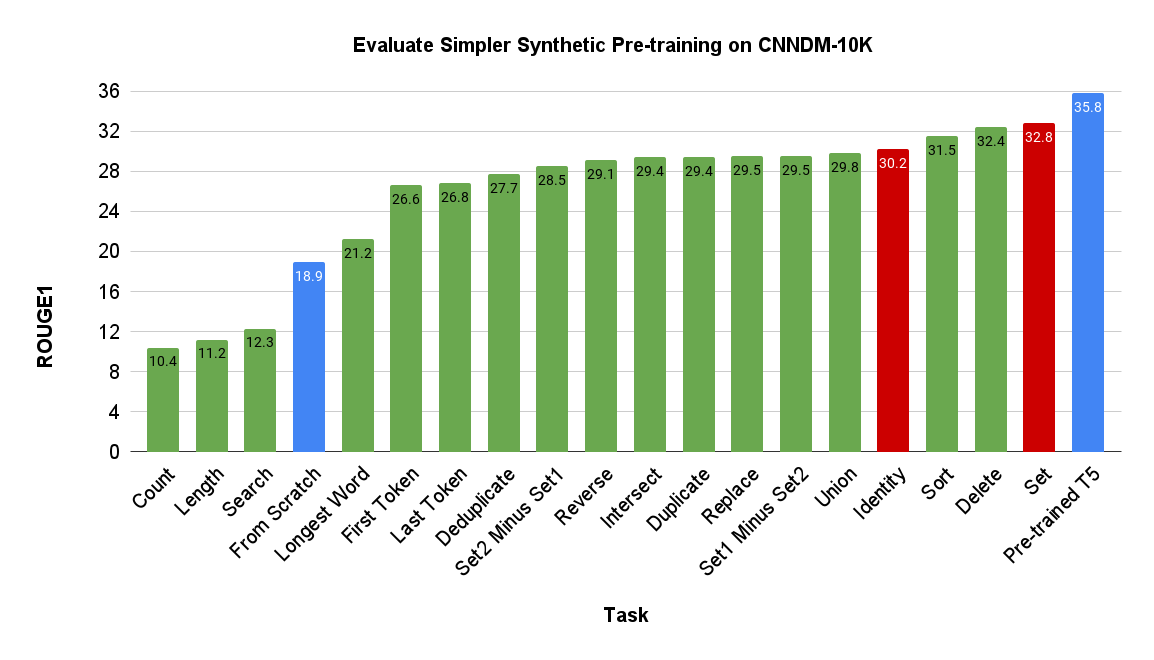}}
\vspace{-0.1in}
\caption{For each simple synthetic task, we pre-train on it and then fine-tune on CNNDM-10K. \set{} performed the best and \identity{}, one of the simplest tasks, provided significant benefit. The \set{} and \identity{} tasks are defined in \ref{section4:simpler_synthetic_tasks}, and the other tasks are defined in \ref{appendix:simpler_synthetic_tasks}.}
\vspace{-0.1in}
\label{section4:figure_simpler_tasks_cnndm10k}
\end{figure}

Encouraged by the results in the previous section, we want to explore what is necessary for synthetic pre-training benefits.
We evaluate pre-training with simpler and more generic synthetic tasks similar to basic Python API functions. Surprisingly, a simple and generic task defined by the \set{} function provides significant benefits ($65\%$), almost matching the previous best synthetic task \lime{} ($67\%$). Furthermore, another simple task defined by the \identity{} function also provided significant benefits ($46\%$). 
\subsection{Simpler Synthetic Tasks}
\label{section4:simpler_synthetic_tasks}
We constructed 18 simpler synthetic tasks similar to basic Python API functions. As we did for the three existing synthetic tasks, for each simpler synthetic task we generated one million data points with a vocabulary of 32K tokens, matching the vocabulary size of \pretrained{}. 
In Appendix~\ref{appendix:simpler_synthetic_tasks_set}, we define each task, explain how examples are randomly generated, and provide input-output examples. We first evaluated all tasks on CNNDM-10K. We then evaluated \set{}, one of the best performing tasks on CNNDM-10K, and \identity{}, one of the simplest tasks, on all six downstream tasks.
\paragraph{Set} 
The \set{} task consists of outputting the input sequence in its original order except without tokens appearing more than once. An example input is $[$a b b a a a c d c$]$, and the corresponding output is $[$a b c d$]$. The \set{} task data generation is explained in Appendix~\ref{appendix:simpler_synthetic_tasks}.
\paragraph{Identity} 
The \identity{} task consists of copying the input tokens to the output. An example input is $[$a d e a a$]$, and the corresponding output is $[$a d e a a$]$. To generate one example, we uniformly sample a sequence of tokens between length 10 and length 220 to be the input and output.

\subsection{Results: simpler synthetic tasks provide similar benefits as existing synthetic tasks}
Figure~\ref{section4:figure_simpler_tasks_cnndm10k} shows results on CNNDM-10K for models pre-trained on each of $18$ simpler synthetic tasks. Table~\ref{section3:table_main} shows results on all six downstream tasks for \set{} and \identity{} pre-trained models. Averaged across the six downstream tasks, \set{} and \identity{} gain $65\%$ and $46\%$ of the natural pre-training benefits compared to $67\%$ for \lime{}, the best previously proposed synthetic task evaluated in Section~\ref{section3}.
\identity{} provides at least as much benefit as Nested Language and Nonsense Summary for all tasks besides Code Translation, for which Nested Language benefits more.

The significant benefits of \set{} and \identity{} pre-training suggest that the complexities of the three previously proposed synthetic tasks we evaluated---representing three fundamental logical operations, having a structural property that mimics natural language, and reflecting operations used for summarization---may not be the necessary factors for why pre-training on those tasks provides benefits. Also, due to their simplicity, \set{} and \identity{} may be useful for future work towards understanding pre-training and how to design better synthetic tasks.

\section{Even Simpler Synthetic Pre-training with Statistics}
\label{section5}

\begin{figure}[t]
\centering 
\centerline{\includegraphics[width=1.0\columnwidth]{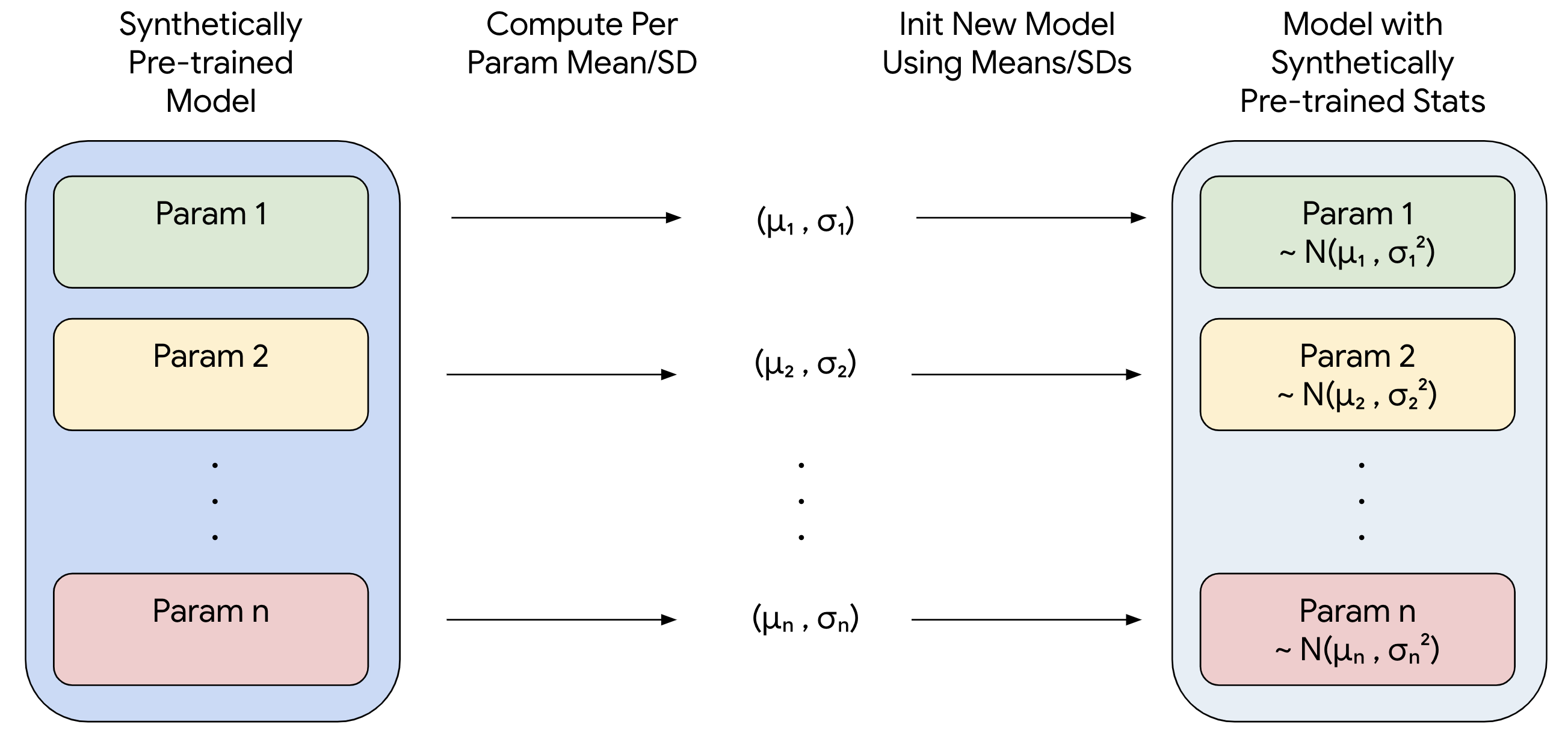}}
\caption{Illustration of Section~\ref{section5:per_param_mean_sd} \perparammeansd{} init scheme. We use this init scheme with \lime{}, \set{}, or \identity{} pre-trained models before fine-tuning on the downstream tasks. As an example, if we were to use this init scheme with an \identity{} pre-trained model, we would initialize the encoder layer 2 query matrix weights by sampling from a Gaussian with the the mean and SD of the encoder layer 2 query matrix from an \identity{} pre-trained model, and we would initialize every other parameter in the same manner. The \perparammeansd{} init scheme recovered more than $70\%$ of the benefit of normally fine-tuning a synthetic pre-trained model (or an average of $39\%$ of natural pre-training benefits).} 
\label{section5:init_figure}
\vspace{-0.15in}
\end{figure}
Inspired by the previous section, we tried to find even simpler initializations that achieve the the benefits of pre-training. We explored how benefits of synthetic pre-training can be attained by initialization based only on statistics of pre-trained weights. 
We found:
(1) Initialization using the mean and standard deviation (SD) of synthetic pre-trained parameters can gain more than $70\%$ of the benefit from synthetic pre-training for CNNDM-10K and MTOP (or an average of $39\%$ of natural pre-training benefits).
    (2) Initialization using simpler schemes involving less statistics can still provide benefits. 
    (3) Specifically the pre-attention layer norm plays an important role for initialization benefits.

\subsection{How much of synthetic pre-training benefit can be attributed to better statistics for initialization?}
\label{section5:per_param_mean_sd}
We compare normal fine-tuning, which we call \fullinit{}, with fine-tuning a model that is initialized with the mean and standard deviation (SD) of each synthetic pre-trained model's parameter. We call this initialization scheme \perparammeansd{} and illustrate it in Figure~\ref{section5:init_figure}. Specifically, for each parameter $p$
(e.g. an attention query matrix) in the synthetic pre-trained model, we compute the the mean $\mu_p$ and the standard deviation $\sigma_p$ of all entries in $p$. We then randomly initialize the model with the same $\mu_p$ and $\sigma_p$ for each parameter $p$. For example, if a pre-trained model had a $2\times 2$ parameter matrix $p = [[-1,2],[2,5]]$, then $\mu_p=2$ and $\sigma_p=2.4$, and we would initialize $p$ by sampling the each entry of this $2\times 2$ matrix from a normal distribution $\mathcal{N}(2, 2.4^2)$. We list all all the parameters we initialize for T5-Small in this manner in Appendix~\ref{appendix:t5_small_parameters_list}.

As shown in Table~\ref{section5:table_main}, using the \perparammeansd{} initialization scheme achieves a large proportion of the benefit from normally fine-tuning a synthetic pre-trained model. For example, for \lime{} \identity{} and \set{}, the \perparammeansd{} initialization recovered $78\%$, $85\%$, and $97\%$\footnote{These percentages are computed by $\frac{\text{performance(\perparammeansd{})} - \text{performance(\randominit{})}}{\text{performance(\fullinit{})}-\text{performance(\randominit{})}}$} of MTOP \fullinit{} performance (or 65\%, 66\%, 69\% of the benefits by natural pre-training). The remaining gap between \perparammeansd{} and \fullinit{} supports the possibility that other factors besides better initialization statistics may cause the benefits of synthetic pre-training. We note that it does not preclude the possibility that better initialization statistics is the sole cause of benefits: there may exist other initialization schemes such that the statistics of the pre-trained model are a subset of the statistics from the initialization scheme.

\begin{table}[]
\caption{Evaluating different initialization schemes with statistics of \lime{}, \set{}, and \identity{} pre-trained models. Results for \perparammeansd{} for the four other tasks as well as five seeds for CNNDM-10K, MTOP, and SQuAD are in Appendix Table~\ref{appendix:table_per_param_mean_sd}. Results for two other initialization ablations and three other \subsetperparamscale{} subset choices we tried are in Appendix Table~\ref{appendix:table_init_main}.
} 
\label{section5:table_main}
\begin{adjustbox}{width=1.0\linewidth}

\begin{tabular}{@{}cccccccc@{}}
                   & \multicolumn{3}{c}{CNNDM-10K}  &  & \multicolumn{3}{c}{MTOP}             \\
                   & \lime{}         & \set{}      & \identity{} &  & \lime{}         & \set{}       & \identity{}      \\ \midrule
\multicolumn{8}{c}{Section~\ref{section5:per_param_mean_sd}}                                                                   \\ \midrule
\fullinit{}   & 33.2         & 32.8     & 30.2 &  & 73.7/94.0    & 71.7/93.7 & 68.6/93.0 \\
\perparammeansd{} & 29.3         & 29.8     & 28.8     &  & 66.1/92.3    & 66.6/92.3 & 67.8/92.6 \\ \midrule
\multicolumn{8}{c}{Section~\ref{section5:simpler_init_schemes}}                                                                   \\ \midrule
\perparamscale{}    & 29.0           & 29.5     & 29.0   &  & 67.3/92.5    & 66.5/92.4 & 66.3/92.3 \\
\wholemodelscale{}  & 15.5         & 26.1     & 20.5 &  & 34.3/81.4    & 38.7/83.1 & 59.8/91.0 \\ 
\preattnlnsubsetperparamscale{} & 27.8 & 25.1 & 27.3 &  & 59.9/90.3 & 58.9/89.0 & 46.7/86.3 \\ 
 &
  \multicolumn{1}{l}{} &
  \multicolumn{1}{l}{} &
  \multicolumn{1}{l}{} &
  \multicolumn{1}{l}{} &
  \multicolumn{1}{l}{} &
  \multicolumn{1}{l}{} &
  \multicolumn{1}{l}{} \\ \midrule
  
Baseline     & \multicolumn{3}{c}{CNNDM-10K}      &                      & \multicolumn{3}{c}{MTOP}     \\ \midrule                                
\randominit{}      & \multicolumn{3}{c}{18.9}      &                      & \multicolumn{3}{c}{38.6/83.0}     \\
\pretrained{}         & \multicolumn{3}{c}{35.8}      & \multicolumn{1}{l}{} & \multicolumn{3}{c}{81.0/95.2}     \\   \midrule       
\end{tabular}
\end{adjustbox}
\end{table}

\subsection{Can simpler initialization schemes still produce benefits?}
\label{section5:simpler_init_schemes}
Next we explore if simpler initialization schemes that use a smaller number of statistics than \perparammeansd{} can still provide benefits. We present three initialization schemes: \perparamscale{}, \wholemodelscale{}, and \subsetperparamscale{}, with results shown in Table~\ref{section5:table_main}. 
We also tried two other initialization schemes, \texttt{Across\_Layers\_Scale} and \texttt{Per\_Layer\_Scale}, discussed in Appendix~\ref{appendix:init_study}. 

\paragraph{Per Param Scale} We use the term ``scale statistic’’ to denote the mean of layer norm parameters or the SD of non layer norm parameters. Unlike the \perparammeansd{} init scheme in the previous subsection, where we initialize each parameter with two statistics (mean and SD), \perparamscale{} initializes each parameter with only the scale statistic: initialize layer norm parameters by setting their value to the mean of the corresponding pre-trained parameter weights, and initialize non layer norm parameters by sampling from a Gaussian with mean 0 and the SD of the corresponding pre-trained parameter weights.

\paragraph{Whole Model Scale}
We initialize the whole model with only two statistics: the SD over all non layer norm weights $\sigma_m$ and the mean over all layer norm weights $\mu_m$. Initialize each non layer norm weight by sampling from $\mathcal{N}(0,\sigma_m^2)$ and Initialize each layer norm weight as the value $\mu_m$.

\paragraph{Per Param Scale: Subset}
The motivation for this init scheme is to find a smaller set of parameters whose initialization can provide significant benefit. We initialize a subset of the parameters (e.g., attention parameters) with the \perparamscale{} initialization and initialize the rest of the parameters with the default \randominit{}. With this initialization scheme, we tried four different subsets: attention parameters, MLP parameters, pre-attention layer norms, and pre-MLP layer norms.

\paragraph{Results} 

    \perparamscale{} provided almost identical benefits as the \perparammeansd{} initialization, suggesting that just the scale statistics, rather than both the mean and SD, are sufficient to provide benefits of initialization.
    
    \wholemodelscale{} initialization was much worse than \perparammeansd{}, yielding almost no improvements. A finer scale initialization rather than simply increasing the scale of weights across the whole model as \wholemodelscale{} does may be necessary for benefits. 
    
    For the \subsetperparamscale{} init scheme, pre-attention layer norms was the only subset we tried that provided significant initialization benefits, covering $46\%$ and $39\%$ of the gap between From Scratch and \pretrained{} for CNNDM-10K and MTOP.  See results in Table~\ref{section5:table_main} (labeled \preattnlnsubsetperparamscale{}) and results for all four subsets tried in Appendix Table~\ref{appendix:table_init_main}.

\begin{table}[]
\caption{Results from initializing pre-attention layer norm parameters to a particular value. They are initialized to the value of $1.0$ in the \randominit{} baseline which is the default T5 initialization.}
\label{section5:table_pre_attn_ln_sweep}
\begin{adjustbox}{width=1.0\linewidth}
\begin{tabular}{@{}ccccccc@{}}
\toprule
Pre-attn LN Init Value & CNNDM-10K & MTOP      & WebQSP    & SQuAD     & Code Trans. & Retrosyn. \\ \midrule
0.05                   & 28.2      & 25.7/73.3 & 30.5/71.2 & 22.8/31.2      & 57.9        & 39.3      \\
0.1                    & 28.3      & 32.5/77.5 & 28.9/70.6 & 23.5/31.7      & 57.6        & 39.9      \\
0.2                    & 28.2      & 56.4/89.3 & 30.6/72.3 & 28.2/37.3      & 57.0        & 39.4      \\
0.4                    & 24.4      & 56.9/89.9 & 30.0/72.8 & 36.3/46.9      & 57.3        & 39.6      \\
0.8                    & 18.6      & 50.7/88.5 & 28.0/71.6 & 34.2/44.6      & 57.2        & 39.9      \\
                       &           &           &           &           &             &           \\ \midrule
Baseline               & CNNDM-10K & MTOP      & WebQSP    & SQuAD     & Code Trans. & Retrosyn. \\ \midrule
\randominit{}            & 18.9 (0.5)      & 38.6/83.0 (2.1/1.2) & 25.8/71.5 (0.7/0.5)      & 16.0/24.2 (2.1/2.0) & 57.6 (0.3)        & 39.0 (0.5)      \\
\pretrained{}         & 35.8 (0.1)      & 81.0/95.2 (0.3/2.2) & 82.3/91.7 (0.8/0.4)      & 77.4/86.1 (0.2/0.2) & 61.2 (0.7)        & 43.5 (0.3)      \\ \bottomrule
\end{tabular}
\end{adjustbox}
\end{table}
\subsection{A Better Pre-attention Layer Norm Initialization}
\label{section5:pre_attention_ln}
\paragraph{Synthetic pre-training results in lower pre-attention layer norm means.}
Motivated by benefits from the \subsetperparamscale{} initialization scheme---initializing pre-attention layer norms using their synthetic pre-trained means and all other parameters with the default initialization---we wanted to see if the pre-attention layer norm mean values in synthetic pre-trained models have a pattern. In Appendix Figure~\ref{appendix:figure_plot_pre_attn_ln}, we plotted pre-attention layer norm parameter means of each layer for \lime{}, \set{}, and \identity{} pre-trained models. We noticed that the pre-attention layer norm means from the synthetic pre-trained models are much smaller values (below $0.4$ for most layers) compared to the default initialization of $1.0$. 

\paragraph{Lower value pre-attention layer norm initialization provides benefits.} Inspired by the above observation, we tried an initialization scheme where all the pre-attention layer norm parameters are initialized to a specific value, and all other parameters are initialized with the default initialization. We tried five different values ranging from $0.05$ to $0.8$. Results in Table~\ref{section5:table_pre_attn_ln_sweep} show that initializing the pre-attention layer norm parameters to a lower value than the default initialization of $1.0$ provided significant benefits: initializing them to $0.2$ improved CNNDM-10K from $18.9$ to $28.2$ ROUGE1, MTOP from $38.6\%$ to $56.4\%$ EM, and SQuAD from 16.0 to 22.8 EM.

\section{Related Work}

\paragraph{Synthetic Pre-training} Past work has shown that synthetic pre-training can benefit a wide range of downstream tasks. For example, it has been shown that pre-training on artificial languages that intuitively mimic properties of natural language can benefit language modeling, dependency parsing, and GLUE~\citep{ri, isabel, chiang}. Pre-training on LIME, a mixture of three synthetic tasks motivated from reasoning primitives, benefits mathematical reasoning benchmarks~\citep{lime}. For summarization, \citet{nonsense_summary} designed a suite of synthetic summarization tasks on a corpus of random tokens, and they show such pre-training can almost match up with natural pre-training. Synthetic pre-training also has been applied outside of text-based domains:~\citet{fractal} shows pre-training on synthetically generated fractal images nearly matches the benefits of pre-training with ImageNet.

\paragraph{Cross-domain Transfer Learning}
Synthetic pre-training benefiting downstream tasks across a wide range of domains is an instance of cross-domain transfer learning, which past work has demonstrated the efficacy of. Pre-training language models with many languages benefits downstream tasks in a new unseen language~\citep{cross_lang_transfer}. More surprisingly, pre-training on music, code, and amino acid sequences benefits natural language downstream tasks~\citep{isabel, chiang_old}. It was also shown recently that pre-training on natural language benefits offline RL~\citep{wiki_rl}.  

\paragraph{Understanding Pre-training} Past work has explored what properties of the pre-training data are important for benefits. \citet{ri} and \citet{isabel} use synthetic pre-training to identify general properties of pre-training data that correlate with improved downstream language task performance by leveraging how synthetic data is easy to modify in a controlled manner. \citet{isabel} found that changing the distribution of tokens in an artificial language can affect downstream performance, and \citet{ri} found that for an artificial brackets language using a nesting structure as well as different tokens for bracket pairs results in better performance than using a flat structure and the same token for bracket pairs. 
\citet{understand_cross_lang_trans_1}, \citet{understand_cross_lang_trans_2}, and \citet{understand_cross_lang_trans_3} explore data properties related to cross-language transfer learning. \citet{word_order} shows that word order within sentences is not essential for natural langauge pre-training benefits.

Other past work has used probing methods to explore what potentially transferable knowledge is gained from pre-training. \citet{ri} explored how much knowledge of "position-aware context dependence" was gained from artificial language and natural language pre-training. \citet{probe_cross_lang_1} and \citet{probe_cross_lang_2} explored potentially transferable knowledge gained from multi-lingual language pre-training.

\paragraph{Initialization} 
Our work tried to understand pre-training through the lens of initialization statistics that benefit optimization. There has been past work on improving transformer initialization to achieve more stable training and benefits with deeper models~\citep{init_4, init_3, init_2, init_1}. These papers do not explore benefits from initializing with pre-trained model statistics.

\section{Discussion}
\label{discussion}

\paragraph{Limitations of current synthetic tasks} In this work, we show that synthetic pre-training can provide significant benefits across many different tasks. However, synthetic pre-training still lags behind natural pre-training by an average of $33\%$ for the six downstream tasks we fine-tuned on. This is expected for natural language downstream tasks, but it is important to note that even on non-natural language tasks, natural language pre-training still outperforms synthetic pre-training. This result suggests that natural language pre-training produces effects that are missing from the evaluated synthetic pre-training. Exploring the differences between different pre-training may be an interesting direction for future work. 

\paragraph{Synthetic pre-training requires deeper understanding} Our results show that initialization using the statistics from synthetic pre-trained models can recover more than $70\%$ of the benefit from synthetic pre-training. The remaining gap leaves open the possibility that the benefits of synthetic pre-training are due to other factors besides better statistics for initialization. Due to their simplicity, \set{} and \identity{} may be useful for future work towards deeper understanding of synthetic pre-training.

In the long term, as we learn more about synthetic pre-training and pre-training in general, we believe that for some downstream tasks it may be possible to develop synthetic pre-training that outperforms natural pre-training: the complexity of existing natural data is fixed, while in some sense the complexity of fully synthetically generated data is infinite. 

\paragraph{Privacy and Ethics} 
This work focuses on the understanding of synthetic pre-training. But synthetic pre-training can also have practical implications for privacy and ethics. 
Standard pre-training data, even when its public, can contain private user information. Large pre-trained models are capable of memorizing training examples, which makes them vulnerable to privacy attacks~\citep{DBLP:journals/corr/abs-2012-07805}.
Synthetic data on the other hand is divorced from the real world, which can help mitigate privacy issues.  Large pre-trained corpora can contain harmful text (e.g., hate speech).  With synthetic pre-training, since we have complete control over what goes into the data, we can potentially mitigate some of the harmful behavior of existing models.
Unfortunately, synthetic pre-training is currently not as performant as natural pre-training, so real-world data would still have to be used to close the performance gap.

\section*{Acknowledgements}
We thank Google TPU Research Cloud for the experimental support. We thank Christian Szegedy, Eric Zelikman, Tri Dao, Dan Fu, Sidd Karamcheti and Rishi Bommasani, for their useful feedbacks on the draft.

\bibliographystyle{neurips_2022}
\bibliography{neurips_2022}
\newpage
\appendix

\appendix

~\section{Synthetic Pre-training Details}
\label{appendix:pretrain_details}
We do synthetic pre-training with the same hyperparameters that the off-the-shelf pre-trained T5-small was trained with: AdaFactor optimizer, batch size 128, sequence length 512, and inverse square root learning rate $1/\sqrt{max(n, 10000)}$ where n is the current step. We evaluate token validation accuracy and save a checkpoint every 5000 steps. For synthetic tasks besides the artificial language, we fine-tuned with the first checkpoint that the model reaches above 99\% token validation accuracy. For LIME, Nonsense Summary, Set, and Identity this was 30K steps, 5K steps, 10K steps, and 5K steps respectively. For the artificial language pre-training, after training for 300K steps the model's token validation accuracy had plateaued without converging to above 99\% token validation accuracy, so we chose to fine-tune with the max accuracy checkpoint, which was at step 165K with accuracy of 77.7\%. For reference, off-the-shelf pre-trained T5-small was trained for 524,288 steps \citep{t5}.
\section{Fine-tuning Details}
\label{appendix:finetune_details}
We fine-tuned with the same hyperparameters \citep{t5} fine-tuned with, with the exception of learning rate, for which we do a sweep over {1e-2, 3e-3, 1e-3}. We use the AdaFactor optimizer and batch size 128 for all except WebQSP and Code Translation, for which we use batch size 32 due to their long sequence length. We list the learning rate and sequence length for each task in Appendix~\ref{appendix:downstream_tasks_hyperparams}. We use the default T5 tokenization \citep{t5}. We fine-tuned for a max of 150K steps for code translation, 400K steps for retrosynthesis, 250K steps for SQuAD, 50K steps for WebQSP, 100K steps for MTOP, and 100K steps for CNNDM-10K. We  evaluate and save checkpoints every 10K steps for all tasks besides WebQSP, for which we evaluate and save checkpoints every 5K steps. For all tasks besides CNNDM-10K and SQuAD, we report the test set metric achieved with the checkpoint corresponding to the max token validation accuracy checkpoint. For CNNDM-10K, we reported the test set metric achieved at step 50K, because we observed the test set metric achieved with the max token validation accuracy checkpoint was often significantly worse than later step checkpoints, and the test set metric always plateaued before and up to about 50K steps. For SQuAD, we report the best validation set metric, as was done in \citep{t5}, because evaluating on the test set requires running inference on a benchmark server.

\section{Off-the-shelf Pre-trained T5v1.1-Small for CNNDM-10K}
\label{appendix:T5v1.1}
The differences between Pre-trained T5v1.1 and Pre-trained T5 are:
\begin{enumerate}
    \item T5v1.1 Uses the GEGLU activation in feed-forward hidden layers rather than ReLU.
    \item Dropout was turned off in pre-training for T5v1.1.
    \item No parameter sharing between the embedding and classifier layer for T5v1.1.
    \item \textbf{T5v1.1 was pre-trained on C4 only, without mixing in the downstream tasks.}
\end{enumerate}
The last difference is why we fine-tune CNNDM-10K with Pre-trained T5v1.1 instead of Pre-trained T5. Using Pre-trained T5 would be an unfair comparison because it has already trained been on all 290K CNNDM task training data.  

\section{Downstream Tasks}
\label{appendix:downstream_tasks}
\subsection{Task Descriptions}
\textbf{CNNDM-10K} is 10K training examples from the CNNDM benchmark, which consists of news articles from CNN and Daily Mail and summaries of each article \citep{cnndm}. We evaluated on CNNDM-10K instead of the full CNNDM because this was the dataset that \citep{nonsense_summary} evaluated their nonsense summarization synthetic task on, which was one of the synthetic tasks we evaluate in Section~\ref{section3}.

\textbf{MTOP} \citep{mtop} and \textbf{WebQSP} \citep{webqsp} are two benchmarks for semantic parsing, the task of converting a natural language query to a logical form. For these tasks, we use the data from \citep{unifiedskg} that is already processed to be in a Seq2Seq format compatible with T5. MTOP has 17K training examples and WebQSP has 2.7K training examples.

\textbf{SQuAD 1.1} is a reading comprehension dataset consisting of ``questions posed by crowdworkers on a set of Wikipedia articles, where the answer to every question is a segment of text or span from the corresponding reading passage'' \citep{squad}. There are 87K training examples.

\textbf{Java to C\# code translation} is one task within Microsoft's CodeXGLUE: General Language Understanding Evaluation benchmark for Code \citep{codexglue}. The task consists of 10K training examples that were collected from the code of several open-source projects that were originally developed in Java and then ported to C\#. 

\textbf{USPTO-50K} is a dataset of 50K chemical reactions that is a commonly used to benchmark deep learning methods for the chemistry task of single-step retrosynthesis, which is predicting possible reactants when given a product as input. \citep{uspto-50K}. For this task, we use the data from \citep{chemistry_t5} that is already processed to be in a Seq2Seq format compatible with T5. There are 40K training examples. 

\subsection{Task Examples}
\paragraph{CNNDM-10K}
\begin{lstlisting}[breaklines]
Source:
summarize: marouane fellaini and adnan januzaj continue to show the world they are not just teammates but also best mates . the manchester united and belgium duo both posted pictures of themselves out at a restaurant on monday night ahead of their game against newcastle on wednesday . januzaj poses in the middle of fellaini and a friend looking like somebody who failed to receive the memo about it being a jackson 5 themed night . premier league duo adnan januzaj and marouane fellaini pose with a friend on the dance floor . manchester united and belgium duo fellaini and januzaj are good friends both on and off the pitch . manchester united ace fellaini runs over to the bench to celebrate his goal against qpr with friend januzaj . the disco effect in the background adds to the theory, but januzaj doesn't seem to mind as they later pose on the dance floor with other friends. united haven't had too many reasons to have a song and dance this season so it seems they may be hitting the discotheques as another form of release . however, victory against newcastle on wednesday would leave manager louis van gaal at least tapping his toes as they continue to fight for a champions league spot this season. januzaj and robin van persie join fellaini in celebrating in front of the manchester united fans at west brom . januzaj receives some words of wisdom from manchester united's dutch manager louis van gaal . januzaj and fellaini are joined by some friends as they take to the dance floor ahead of the newcastle game .

Target:
the belgian duo took to the dance floor on monday night with some friends . manchester united face newcastle in the premier league on wednesday . red devils will be looking for just their second league away win in seven . louis van gaal's side currently sit two points clear of liverpool in fourth .
\end{lstlisting}

\paragraph{MTOP}
\begin{lstlisting}[breaklines]
Source:
Has Angelika Kratzer video messaged me ? ; structured knowledge: IN:GET: MESSAGE, WEATHER, ALARM, INFO_RECIPES, STORIES_NEWS, REMINDER, RECIPES, EVENT, CALL_TIME, LIFE_EVENT, INFO_CONTACT, CONTACT, TIMER, REMINDER_DATE_TIME, AGE, SUNRISE, EMPLOYER, EDUCATION_TIME, JOB, AVAILABILITY, CATEGORY_EVENT, CALL, EMPLOYMENT_TIME, CALL_CONTACT, LOCATION, TRACK_INFO_MUSIC, SUNSET, MUTUAL_FRIENDS, UNDERGRAD, REMINDER_LOCATION, ATTENDEE_EVENT, MESSAGE_CONTACT, REMINDER_AMOUNT, DATE_TIME_EVENT, DETAILS_NEWS, EDUCATION_DEGREE, MAJOR, CONTACT_METHOD, LIFE_EVENT_TIME, LYRICS_MUSIC, AIRQUALITY, LANGUAGE, GENDER, GROUP | IN:SEND: MESSAGE | IN:SET: UNAVAILABLE, RSVP_YES, AVAILABLE, DEFAULT_PROVIDER_MUSIC, RSVP_INTERESTED, DEFAULT_PROVIDER_CALLING, RSVP_NO | IN:DELETE: REMINDER, ALARM, TIMER, PLAYLIST_MUSIC | IN:CREATE: ALARM, REMINDER, CALL, PLAYLIST_MUSIC, TIMER | IN:QUESTION: NEWS, MUSIC | IN:PLAY: MUSIC, MEDIA | IN:END: CALL | IN:IGNORE: CALL | IN:UPDATE: CALL, REMINDER_DATE_TIME, REMINDER_TODO, TIMER, METHOD_CALL, ALARM, REMINDER_LOCATION, REMINDER | IN:PAUSE: MUSIC, TIMER | IN:ANSWER: CALL | IN:SNOOZE: ALARM | IN:IS: TRUE_RECIPES | IN:REMOVE: FROM_PLAYLIST_MUSIC | IN:ADD: TIME_TIMER, TO_PLAYLIST_MUSIC | IN:SHARE: EVENT | IN:PREFER:  | IN:START: SHUFFLE_MUSIC | IN:SILENCE: ALARM | IN:SWITCH: CALL | IN:SUBTRACT: TIME_TIMER | IN:PREVIOUS: TRACK_MUSIC | IN:HOLD: CALL | IN:SKIP: TRACK_MUSIC | IN:LIKE: MUSIC | IN:RESTART: TIMER | IN:RESUME: TIMER, CALL, MUSIC | IN:MERGE: CALL | IN:REPLAY: MUSIC | IN:LOOP: MUSIC | IN:STOP: MUSIC, SHUFFLE_MUSIC | IN:UNLOOP: MUSIC | IN:CANCEL: MESSAGE, CALL | IN:REWIND: MUSIC | IN:REPEAT: ALL_MUSIC, ALL_OFF_MUSIC | IN:FAST: FORWARD_MUSIC | IN:DISLIKE: MUSIC | IN:DISPREFER:  | IN:HELP: REMINDER | IN:FOLLOW: MUSIC     

Target:
[IN:GET_MESSAGE [SL:CONTACT Angelika Kratzer ] [SL:TYPE_CONTENT video ] [SL:RECIPIENT me ] ]
\end{lstlisting}

\paragraph{WebQSP}
\begin{lstlisting}[breaklines]
Source:
where is isthmus of panama located? ; structured knowledge: Isthmus of Panama: m.04zwft | m.06n3y location.location.contains m.06w99sr | m.06n3y location.location.contains m.0cn1d6 | m.06n3y location.location.contains m.0c_ys | m.06n3y location.location.contains m.037l8t | m.06n3y location.location.contains m.0w_hmnw | m.06n3y location.location.contains m.0c4rq4 | m.06n3y location.location.contains m.02t1x0 | m.06n3y location.location.contains m.05w5vr | m.04zwft common.topic.notable_types m.01nt | m.06n3y location.location.contains m.09vgjp | m.06n3y location.location.contains m.0b2ft5 | m.06n3y location.location.contains m.07twz | m.06n3y location.location.contains m.0cqy3y | m.04zwft location.location.containedby m.06n3y | m.06n3y location.location.contains m.02x1y_b | m.06n3y location.location.contains m.03t3qb | m.06n3y location.location.contains m.06w57th | m.06n3y location.location.contains m.06w9q8y | m.06n3y location.location.contains m.0cpbxws | m.06n3y location.location.contains m.07t1sw | m.06n3y location.location.contains m.01nr2h | m.06n3y location.location.contains m.06gldq | m.06n3y location.location.contains m.034m8 | m.06n3y location.location.contains m.0dkz7x | m.06n3y location.location.contains m.0c7r4h | m.06n3y location.location.contains m.025vjdw | m.06n3y location.location.contains m.02qbjjz | m.06n3y location.location.contains m.0wq95z_ | m.06n3y location.location.contains m.0sd7 | m.06n3y location.location.contains m.06bf8s | m.06n3y location.location.contains m.027d7t4 | m.06n3y location.location.contains m.0cn1vs | m.06n3y location.location.contains m.076ycbb | m.06n3y location.location.contains m.061b7x | m.06n3y location.location.contains m.0p2n | m.06n3y location.location.contains m.0w_j2g5 | m.06n3y location.location.contains m.015fr | m.06n3y location.location.contains m.02xsr6 | m.06n3y location.location.contains m.02z722h | m.06n3y location.location.contains m.06w3qq8 | m.06n3y location.location.contains m.0cn1k1 | m.06n3y location.location.contains m.02vrgz0 | m.06n3y location.location.contains m.0g61h6 | m.06n3y location.location.contains m.0w_j7px | m.06n3y location.location.contains m.05fzj6 | m.06n3y location.location.contains m.026h5zg | m.06n3y location.location.contains m.0jgd | m.06n3y location.location.contains m.03qjmgc | m.06n3y location.location.contains m.01ls2 | m.06n3y location.location.containedby m.07c5l 

Target:
(JOIN (R location.location.containedby) m.04zwft)
\end{lstlisting}

\paragraph{SQuAD 1.1}
\begin{lstlisting}[breaklines]
Source:
question: What does increased oxygen concentrations in the patient's lungs displace? context: Hyperbaric (high-pressure) medicine uses special oxygen chambers to increase the partial pressure of O 2 around the patient and, when needed, the medical staff. Carbon monoxide poisoning, gas gangrene, and decompression sickness (the 'bends') are sometimes treated using these devices. Increased O 2 concentration in the lungs helps to displace carbon monoxide from the heme group of hemoglobin. Oxygen gas is poisonous to the anaerobic bacteria that cause gas gangrene, so increasing its partial pressure helps kill them. Decompression sickness occurs in divers who decompress too quickly after a dive, resulting in bubbles of inert gas, mostly nitrogen and helium, forming in their blood.  Increasing the pressure of O 2 as soon as possible is part of the treatment.
  
Target:
carbon monoxide
\end{lstlisting}

\paragraph{Java to C\# Code Translation}
\begin{lstlisting}[breaklines]
Source:
public void delete(int key) {int i = binarySearch(mKeys, 0, mSize, key);if (i >= 0) {if (mValues[i] != DELETED) {mValues[i] = DELETED;mGarbage = true;}}}   

Target:
public virtual void delete(int key){int i = binarySearch(mKeys, 0, mSize, key);if (i >= 0){if (mValues[i] != DELETED){mValues[i] = DELETED;mGarbage = true;}}}
\end{lstlisting}

\paragraph{USPTO-50K Retrosynthesis}
\begin{lstlisting}[breaklines]
Source:
COc1ccc(CN(C(=O)OCc2ccccc2)[C@@H]2C(=O)N(Cc3ccc(OC)cc3OC)[C@@H]2CC=C(Br)Br)cc1

Target:
BrC(Br)(Br)Br.COc1ccc(CN(C(=O)OCc2ccccc2)[C@@H]2C(=O)N(Cc3ccc(OC)cc3OC)[C@@H]2CC=O)cc1
\end{lstlisting}

\subsection{Downstream Task Specific Hyperparameters}
More general fine-tuning hyperparameters and details are in Appendix~\ref{appendix:finetune_details}. For each task, we chose the learning rate by trying {1e-2, 3e-3, 1e-3} for Random Init finetuning, and picking the best performing value.
\label{appendix:downstream_tasks_hyperparams}.
\paragraph{CNNDM-10K}
We use the same input length, taget length, and max decoding length parameters as the Nonsense Summary synthetic task paper \citep{nonsense_summary}. 
\begin{itemize}
    \item learning rate: 0.001
    \item input length: 512
    \item target length: 256
    \item max decode length: 148
\end{itemize}
\paragraph{MTOP}
\begin{itemize}
    \item learning rate: 0.001
    \item input length: 1024
    \item target length: 128
\end{itemize}
\paragraph{WebQSP}
\begin{itemize}
    \item learning rate: 0.001
    \item input length: 2048
    \item target length: 256
\end{itemize}
\paragraph{SQuAD}
\begin{itemize}
    \item learning rate: 0.001
    \item input length: 512
    \item target length: 128
\end{itemize}
\paragraph{Code Translation}\begin{itemize}
    \item learning rate: 0.003
    \item input length: 1024
    \item target length: 1024
\end{itemize}
\paragraph{Retrosynthesis}
\begin{itemize}
    \item learning rate: 0.003
    \item input length: 256
    \item target length: 256
\end{itemize}

\section{Three Previously Proposed Synthetic Tasks from Section~\ref{section3}}
\label{appendix:three_existing_synthetic_task}
\subsection{LIME}
The input token length ranges from 68 to 214 and the output token length ranges from 7 to 74. For the following LIME examples, the ``rule string'' is $A * A + B = C$, the ``case dictionary'' that represents substitutions is $\{A : a, B : b, C : d + e\}$, and the ``result string'' is $a * a + b = d + e$. In real generated data, the "Rule Symbols" and "Math Symbols" are randomly sampled from a vocabulary of 32K tokens for each data point.
\paragraph{LIME: Induct}
\begin{lstlisting}[breaklines]
Source: 
<RuleSymbols> A B C <MathSymbols> * + = a b d e <s> {A : a, B : b, C : d + e} <s> a * a + b = d + e

Target: 
A * A + B = C
\end{lstlisting}

\paragraph{LIME: Abduct}
\begin{lstlisting}[breaklines]
Source: 
<RuleSymbols> A B C <MathSymbols> * + = a b d e <s> A * A + B = C <s> a * a + b = d + e

Target: 
{A : a, B : b, C : d + e}
\end{lstlisting}
\paragraph{LIME: Deduct}
\begin{lstlisting}[breaklines]
Source: 
<RuleSymbols> A B C <MathSymbols> * + = a b d e <s> A * A + B = C <s> {A : a, B : b, C : d + e}

Target: 
a * a + b = d + e
\end{lstlisting}

\subsection{Dyck Artificial Language}
The input token length ranges from 3 to 512 and the output token length ranges from 3 to 114. For clarity, the following example will involve only 3 head and tail tokens represented by the symbols \{, \}, [, ], (, ). In real generated data, there will be 16K possible head-tail token pairs. \\\\
An example sentence: 
\begin{lstlisting}[breaklines]
[ ( { [ ] } [ ] ) ] { [ ] }
\end{lstlisting}
\mbox{}\\
This sentence converted to a Seq2Seq task with the T5 pre-training objective: 
\begin{lstlisting}[breaklines]
Source:
[ <s0> <s1> [ ] } <s2> ] ) <s3> { [ ] <s4>

Target:
( <s0> { <s1> [ <s2> ] <s3> }
\end{lstlisting}

\subsection{Nonsense Summary}
The input token length ranges from 175 to 709 and the output token length ranges from 12 to 825. We provide an example of a nonsense summary data point in Appendix Figure~ \ref{appendix:figure_example_nonsense_summary}, which is directly taken from the task's original paper \citep{nonsense_summary}.

\section{Dataset Size Ablation Full Results}
The table showing the full training dataset size ablation results is in Appendix Table~ \ref{appendix:table_dataset_size_ablation}.

\section{Simpler Synthetic Tasks from Section~\ref{section4}}
All data is generated by sampling over 32K tokens. But for clarity, all examples are written with individual alphabetic characters representing tokens. All simpler synthetic tasks have input between length 10 and length 220. 
\label{appendix:simpler_synthetic_tasks} 
\subsection{Set} 
\label{appendix:simpler_synthetic_tasks_set} 
\paragraph{Definition}
Input is a token sequence. Output is the input token sequence with no duplicates (in original order).
\paragraph{Example}
\begin{lstlisting}[breaklines]
Source: 
a b a a b b c d

Target: 
a b c d
\end{lstlisting}
\paragraph{Input Generation}
To generate one data point, we do not just uniformly sample tokens. Because over a vocabulary of 32K tokens, that simple sampling would result in few sequences with duplicate tokens where the task input and output are different. Instead, we uniformly sample the length $l$ of the input from min length to max length, then sample $t$ tokens that will be in the input where $t<l$, then sample the amount of each token by randomly partitioning $l$ to $t$ parts $\{p_1, p_2... p_t\}$, then finally generate a random input sequence based on the $t$ sampled tokens and corresponding amounts. 

\subsection{Delete} 
\paragraph{Definition}
Input is token sequence 1 and token sequence 2. Output is sequence 1 with the first occurrence of sequence 2 deleted.
\paragraph{Example}
\begin{lstlisting}[breaklines]
Source: 
c b a a b b a d <s> b a

Target: 
b a b b a d
\end{lstlisting}
\paragraph{Input Generation}
With 0.75 probability, generate sequence 1 and sequence 2 such that sequence 2 appears in sequence 1 at least once via the following process: sample sequence 2, sample sequence 1 length, sample location that sequence 2 appears in sequence 1, sample remaining tokens in sequence 1. 
With 0.25 probability, generate sequence 1 and sequence 2 such that sequence 2 does not appear in sequence 1 via the following process: sample sequence 1, sample sequence 2 and keep re-sampling sequence 2 until it does not appear in sequence 1.

\subsection{Sort} 
\paragraph{Definition}
Input is token sequence 1 and token sequence 2. Output is input sequence 1 sorted in according to order specified by token sequence 2. 
\paragraph{Example}
\begin{lstlisting}[breaklines]
Source: 
c b a a b c b a d <s> b a c d

Target: 
b b b a a a c c d
\end{lstlisting}
\paragraph{Input Generation}
To generate token sequence 1, uniformly sample the length $l$ of the input from min length to max length, then sample $t$ tokens that will be in the input where $t<l$, then sample the amount of each token by randomly partitioning $l$ to $t$ parts $\{p_1, p_2... p_t\}$, then finally generate a random input sequence based on the $t$ sampled tokens and corresponding amounts. To generate sequence 2, shuffle the set of tokens in sequence 1.


\subsection{Identity} 
\paragraph{Definition}
Input is a token sequence. Output is the same as the input token sequence.
\paragraph{Example}
\begin{lstlisting}[breaklines]
Source: 
c b a a b 

Target: 
c b a a b
\end{lstlisting}
\paragraph{Input Generation}
Sample input sequence length $l$ uniformly between min and max possible length, then sample $l$ tokens.

\subsection{Union} 
\paragraph{Definition}
Input is token sequence 1 and token sequence 2. Output is token sequence that is the union of the set of tokens in sequence 1 and set of tokens in sequence 2. 
\paragraph{Example}
\begin{lstlisting}[breaklines]
Source: 
c b a a b <s> c b a e f

Target: 
c b a e f
\end{lstlisting}
\paragraph{Input Generation}
Sample max length $l$, then sample partition of $l$ to $l_1$ and $l_2$, which will be the lengths of sequence 1 and sequence 2. Sample $l/2$ unique tokens. To get sequence 1, sample $l_1$ of the unique tokens. To get sequence 2, sample $l_2$ of the unique tokens.

\subsection{Set1 Minus Set2} 
\paragraph{Definition}
Input is token sequence 1 and token sequence 2. Output is a token sequence that is the subtraction of the set of tokens in sequence 2 from the set of tokens in sequence 1. 
\paragraph{Example}
\begin{lstlisting}[breaklines]
Source: 
c b a a b <s> c a e f

Target: 
b
\end{lstlisting}
\paragraph{Input Generation}
Sample max length $l$, then sample partition of $l$ to $l_1$ and $l_2$, which will be the lengths of sequence 1 and sequence 2. Sample $l/2$ unique tokens. To get sequence 1, sample $l_1$ of the unique tokens. To get sequence 2, sample $l_2$ of the unique tokens.

\subsection{Replace} 
\paragraph{Definition}
Input is token sequence 1 and token sequence 2. Token sequence 2 always has two tokens, which we will denote as $t_{1}$ and $t_{2}$. Output is token sequence 1 with $t_{1}$ replaced by $t_{2}$.
\paragraph{Example}
\begin{lstlisting}[breaklines]
Source: 
c b a a b <s> a x

Target: 
c b x x b
\end{lstlisting}
\paragraph{Input Generation}
To get sequence 1, sample input sequence length $l$ uniformly between min and max possible length, then sample $l$ tokens. To get sequence 2, sample one token from the set of tokens in sequence 1, and sample a token from the entire vocab.
 
\subsection{Duplicate} 
\paragraph{Definition}
Input is a token sequence. Output is the input token sequence with every token duplicated. 
\paragraph{Example}
\begin{lstlisting}[breaklines]
Source: 
c a b a a b 

Target: 
c c a a b b a a a a b b 
\end{lstlisting}
\paragraph{Input Generation}
Sample input sequence length $l$ uniformly between min and max possible length, then sample $l$ tokens.

\subsection{Intersect} 
\paragraph{Definition}
Input is token sequence 1 and token sequence 2. Output is token sequence that is the intersection of the set of tokens in sequence 1 and set of tokens in sequence 2. 
\paragraph{Example}
\begin{lstlisting}[breaklines]
Source: 
c a b c a a b <s> a a b

Target: 
a b
\end{lstlisting}
\paragraph{Input Generation}
Sample max length $l$, then sample partition of $l$ to $l_1$ and $l_2$, which will be the lengths of sequence 1 and sequence 2. Sample $l/2$ unique tokens. To get sequence 1, sample $l_1$ of the unique tokens. To get sequence 2, sample $l_2$ of the unique tokens.

\subsection{Reverse} 
\paragraph{Definition}
Input is a token sequence. Output is the input token sequence in reverse order.
\paragraph{Example}
\begin{lstlisting}[breaklines]
Source: 
c a b c a a b 

Target: 
b a a c b a c
\end{lstlisting}
\paragraph{Input Generation}
Sample input sequence length $l$ uniformly between min and max possible length, then sample $l$ tokens.

\subsection{Set2 Minus Set1} 
\paragraph{Definition}
Input is token sequence 1 and token sequence 2. Output is a token sequence that is the subtraction of the set of tokens in sequence 1 from the set of tokens in sequence 2. 
\paragraph{Example}
\begin{lstlisting}[breaklines]
Source: 
c b c b b <s> a a b x y z

Target: 
a x y z
\end{lstlisting}
\paragraph{Input Generation}
Sample max length $l$, then sample partition of $l$ to $l_1$ and $l_2$, which will be the lengths of sequence 1 and sequence 2. Sample $l/2$ unique tokens. To get sequence 1, sample $l_1$ of the unique tokens. To get sequence 2, sample $l_2$ of the unique tokens.

\subsection{Deduplicate} 
\paragraph{Definition}
Input is a token sequence. Output is the input token sequence with no adjacent duplicate tokens.
\paragraph{Example}
\begin{lstlisting}[breaklines]
Source: 
c b b c c d d d d e f e

Target: 
c b c d e f e
\end{lstlisting}
\paragraph{Input Generation}
Uniformly sample the length $l$ of the input from min length to max length, then sample $t$ tokens that will be  in the input where $t<l$, then sample the amount of each token by randomly partitioning $l$ to $t$ parts $\{p_1, p_2... p_t\}$, then generate the input sequence by iterating through the $t$ tokens and repeating each one $p_i$ times.

\subsection{Last Token} 
\paragraph{Definition}
Input is a token sequence. Output is the last token in the input sequence.
\paragraph{Example}
\begin{lstlisting}[breaklines]
Source: 
c b b c c a

Target: 
a
\end{lstlisting}
\paragraph{Input Generation}
Sample input sequence length $l$ uniformly between min and max possible length, then sample $l$ tokens.

\subsection{First Token} 
\paragraph{Definition}
Input is a token sequence. Output is the first token in the input sequence.
\paragraph{Example}
\begin{lstlisting}[breaklines]
Source: 
c b b c c a

Target: 
c
\end{lstlisting}
\paragraph{Input Generation}
Sample input sequence length $l$ uniformly between min and max possible length, then sample $l$ tokens.

\subsection{Longest Word} 
\paragraph{Definition}
Input is a token sequence. The `input token sequence has a special format, with special separator tokens distributed throughout. Output is the largest number of contiguous tokens.
\paragraph{Example}
\begin{lstlisting}[breaklines]
Source: 
c b <s> s <s> b a f <s> s d <s> a a b c w q 
Target: 
6
\end{lstlisting}
\paragraph{Input Generation}
Sample the number of ``words'' $w$ defined as a contiguous block of normal tokens. Sample the length of the input token sequence $l$. Partition $l-w+1$ to $w$ parts $p_1, p_2, ... p_w$. Build the input sequence by sampling the ``words'' according to the word lengths $p_1, p_2, ... p_w$, and then joining the words with the special separator token.

\subsection{Search} 
\paragraph{Definition}
Input is token sequence 1 and token sequence 2. Output is ``yes'' if token sequence 2 is contained in token sequence 1, and ``no'' otherwise. 
\paragraph{Example}
\begin{lstlisting}[breaklines]
Source: 
a a b c w q a b d e <s> c w e

Target: 
no
\end{lstlisting}
\paragraph{Input Generation}
With 0.75 probability, generate sequence 1 and sequence 2 such that sequence 2 appears in sequence 1 at least once by using the following procedure: sample sequence 2, sample sequence 1 length, sample location that sequence 2 appears in sequence 1, sample remaining tokens in sequence 1. 
With 0.25 probability, generate sequence 1 and sequence 2 such that sequence 2 does not appear in sequence 1 by using the following procedure: sample sequence 1, sample sequence 2 and keep re-sampling sequence 2 until it does not appear in sequence 1.

\subsection{Length} 
\paragraph{Definition}
Input is a token sequence. Output is the number of tokens in the input token sequence.
\paragraph{Example}
\begin{lstlisting}[breaklines]
Source: 
a a b c w 

Target: 
5
\end{lstlisting}
\paragraph{Input Generation}
Sample input sequence length $l$ uniformly between min and max possible length, then sample $l$ tokens.

\subsection{Count} 
\paragraph{Definition}
Input is token sequence 1 and token sequence 2. Token sequence 2 is only 1 token which we will denote $t$. Output is the number of times $t$ occurs in token sequence 1.
\paragraph{Example}
\begin{verbatim}
Source: 
a a b c a a a w <s> a

Target: 
5
\end{verbatim}
\paragraph{Input Generation}
Start by sampling the single token $t$ in sequence 2. Then sample $l$, which will be the length of sequence 1, and sample the number of times $c$ that $t$ will appear in token 1. Sample $l-c$ remaining tokens.




\section{More Initialization Schemes}
\label{appendix:init_study}

We define two other initialization schemes we tried, and show results in Appendix Table~\ref{appendix:table_init_main}
\paragraph{Across Layers Scale}
We tried the \acrosslayers{} initialization scheme, where we initialized each \textit{type} of parameter with the scale statistic of it computed across all layers. Some examples of types of parameters include query matrices, pre-MLP layer norms, and pre-attention layer norms. Using this initialization, all query matrices across the whole model would be initialized with a single SD, which is the SD over all the pre-trained Q matrix weights, across all layers. We find that \acrosslayers{} provides significant benefits for \lime{}, \identity{}, and \set{}, preserving at least 87\% of the Per Param Mean/SD initialization scheme benefit for CNNDM-10K and MTOP. This result suggests that parameter statistics varying from layer to layer might not be an important factor for synthetic pre-training initialization benefits. 

\paragraph{Per Layer Scale}
We tried the \perlayer{} initialization scheme, where every layer is initialized with just two statistics: the SD of all the non-layer norm parameter weights in that layer and the mean of all the layer norm parameter weights in that layer. Results are mixed. There is no benefit for \identity{} CNNDM-10K and minimal benefit for \lime{} MTOP. For the other four pre-training and downstream task settings, there was significant benefit, preserving at least 80\% of the Per Param Mean/SD benefit for CNNDM-10K and MTOP.

\section{Compute Used}
\label{appendix:compute}
Each experiment was ran on a Google Cloud v3-8 TPU VM. We ran most experiments using a total of 5 such VMs.

\section{Additional Experiments}
\paragraph{Multiple Seeds for Section~\ref{section3} and \ref{section4} Experiments}
In Appendix Table~\ref{appendix:table_main_with_std}, we show the mean and standard deviation for five seeds from evaluating Pre-trained T5, Random Init, LIME, and Set on the six downstream tasks. For LIME and Set, for each different seed, we generated and pre-trained on a new synthetic dataset. 

\paragraph{More Tasks for Section~\ref{section5:per_param_mean_sd} Per Param Mean/SD Initialization}
In Appendix Table~\ref{appendix:table_per_param_mean_sd}, we show results for the Section~\ref{section5:per_param_mean_sd} Per Param Mean/SD init scheme for all six tasks, not just CNNDM-10K and MTOP 
which were the only two tasks shown in Table~\ref{section5:table_main}. 
\paragraph{Multiple Seeds for Section~\ref{section5:per_param_mean_sd} Per Param Mean/SD Initialization}
On CNNDM-10K, MTOP, and SQuAD, we run the Per Param Mean/SD ininitialization experiments for five seeds. Results are shown in Appendix Table~\ref{appendix:table_per_param_mean_sd}. We initialized using a model pre-trained on a different generated synthetic dataset for each seed.

\paragraph{More Tasks for Section~\ref{section5:pre_attention_ln} Pre-attention Layer Norm Initialization Sweep}
In Appendix Table~\ref{appendix:table_pre_attn_ln_sweep}, we show results for the Section~\ref{section5:pre_attention_ln} pre-attention layer norm initialization sweep for all six tasks, not just CNNDM-10K and MTOP which were the only two tasks shown in Table~\ref{section5:table_pre_attn_ln_sweep}. 

\newpage

\begin{table}[h]
\caption{Dataset size ablation. Retrosynthesis-40K and SQuAD-87K are the same as Retrosynthesis and SQuAD in Table~\ref{section3:table_main}. The CNNDM-10K and CNNDM-100K results as well as the Retrosynthesis-10K and Reotrosynthesis-40K results show that benefits from synthetic (and T5 natural language) pre-training increase as the amount of training data for a task decrease.  }
  \label{appendix:table_dataset_size_ablation}
\centering
\begin{tabular}{@{}cccc@{}}
\toprule
\multicolumn{1}{l}{} & Retrosyn-1K & Retrosyn-10K & Retrosyn-40K \\ \midrule
Pre-trained T5    & 7.6                  & 29.5                 & 43.1                 \\
LIME        & 4.0                    & 25.1                 & 41.1                 \\
Random Init     & 0.0                    & 18                   & 39.2                 \\
\multicolumn{1}{l}{} & \multicolumn{1}{l}{} & \multicolumn{1}{l}{} & \multicolumn{1}{l}{} \\ \midrule
\multicolumn{1}{l}{} & CNNDM-1K   & CNNDM-10K   & CNNDM-100K  \\ \midrule
Pre-trained T5    & 29.6               & 35.8                 &   39.1               \\
LIME        & 28.4                 & 33.2                 & 37.6                 \\
Random Init     & 12.2                 & 18.9                 & 35.1                 \\
\multicolumn{1}{l}{} & \multicolumn{1}{l}{} & \multicolumn{1}{l}{} & \multicolumn{1}{l}{} \\ \midrule
\multicolumn{1}{l}{} & SQuAD-1K    & SQuAD-10K   & SQuAD-87K   \\ \midrule
Pre-trained T5    & 19.5/36.8          & 59.2/73.5               & 77.5/86.0 \\
LIME       & 3.7/8.5              & 23.21/34.1           & 50.4/62.1            \\
Random Init     & 0.8/0.9              & 0.3/1.6              & 17.2/25.2            \\
\multicolumn{1}{l}{}
\end{tabular}
\end{table}


\begin{table}[h]
\caption{Evaluating different initialization schemes with statistics of LIME, Set, and Identity pre-trained models. } 
\begin{adjustbox}{center}
\label{appendix:table_init_main}

\begin{tabular}{@{}cccccccc@{}}
                   & \multicolumn{3}{c}{CNNDM-10K}  &  & \multicolumn{3}{c}{MTOP}             \\
                   & LIME         & Set      & Identity &  & LIME         & Set       & Identity      \\ \midrule
\multicolumn{8}{c}{Section~\ref{section5:per_param_mean_sd}}                                                                   \\ \midrule
Full Init   & 33.2         & 32.8     & 30.2 &  & 73.7/94.0    & 71.7/93.7 & 68.6/93.0 \\
Per Param Mean/SD & 29.3         & 29.8     & 28.8     &  & 66.1/92.3    & 66.6/92.3 & 67.8/92.6 \\ \midrule
\multicolumn{8}{c}{Section~\ref{section5:simpler_init_schemes}}                                                                   \\ \midrule
Per Param Scale    & 29.0           & 29.5     & 29.0   &  & 67.3/92.5    & 66.5/92.4 & 66.3/92.3 \\
\textbf{Across Layers Scale} & 27.8         & 28.8     & 28.5 &  & 63.5/91.5    & 66.9/92.5 & 67.5/92.6 \\
\textbf{Per Layer Scale}    & 27.1         & 28.1     & 19.6 &  & 48.1/85.3    & 61.1/90.5 & 60.9/90.5 \\
Whole Model Scale  & 15.5         & 26.1     & 20.5 &  & 34.3/81.4    & 38.7/83.1 & 59.8/91.0 \\ \midrule
Pre-attn LN Per Param Scale & 27.8 & 25.1 & 27.3 &  & 59.9/90.3 & 58.9/89.0 & 46.7/86.3 \\ 
\textbf{Attention Per Param Scale}   & 17.2 & 19.1 & 19.1 && 19.3/73.9 & 41.5/84.9 & 38.4/82.9 \\
\textbf{Pre-MLP LN Per Param Scale}  & 20.2 & 20.6 & 20.5 && 41.8/85.1 & 40.3/84.5 & 43.0/85.4   \\
\textbf{MLP Per Param Scale}         & 16.6 & 18.7 & 18.3 && 36.5/80.8 & 36.5/81.6 & 37.9/83.0 \\
 &
  \multicolumn{1}{l}{} &
  \multicolumn{1}{l}{} &
  \multicolumn{1}{l}{} &
  \multicolumn{1}{l}{} &
  \multicolumn{1}{l}{} &
  \multicolumn{1}{l}{} &
  \multicolumn{1}{l}{} \\ \midrule
  
Baseline     & \multicolumn{3}{c}{CNNDM-10K}      &                      & \multicolumn{3}{c}{MTOP}     \\ \midrule                                
Random Init      & \multicolumn{3}{c}{18.9}      &                      & \multicolumn{3}{c}{38.6/83.0}     \\
Pre-trained T5         & \multicolumn{3}{c}{35.8}      & \multicolumn{1}{l}{} & \multicolumn{3}{c}{81.0/95.2}     \\       
\end{tabular}
\end{adjustbox}
\end{table}

\newpage

\begin{figure}[h]
\centering 
\centerline{\includegraphics[width=0.5\columnwidth]{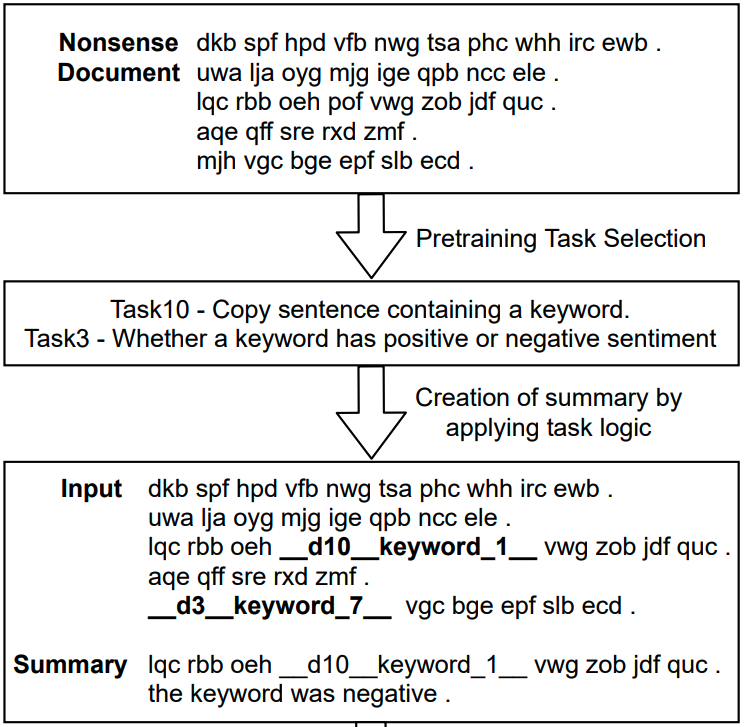}}
\caption{Example of a Nonsense Summary task data point. Figure is directly taken from \citep{nonsense_summary}.}
\label{appendix:figure_example_nonsense_summary}
\vspace{-0.1in}
\end{figure}



\begin{table}[h]
\caption{Same table as Table~\ref{section3:table_main} in main paper except with mean and standard deviation over 5 seeds for Pre-trained T5, Random Init, LIME, and Set. For LIME and Set, for each seed, a different synthetic pre-training set was generated and pre-trained on.}
  \label{appendix:table_main_with_std}
\begin{adjustbox}{width=1.0\linewidth}
\begin{tabular}{@{}ccccccc@{}}
                & CNNDM-10K & MTOP      & WebQSP    & SQuAD     & Code Trans. & Retrosyn. \\ 
                & ROUGE1   & EM/F1     & EM/F1     & EM/F1     & EM          & EM        \\ \midrule
\multicolumn{7}{c}{Baselines}                                                            \\ \midrule
Pre-trained T5       & \textbf{35.8 (0.1)}     & \textbf{81.0/95.2 (0.3/2.2)} & \textbf{82.3/91.7 (0.8/0.4)} & \textbf{77.4/86.1 (0.2/0.2)} & \textbf{61.2 (0.7)}        & \textbf{43.5 (0.3)}      \\
Wiki 10K        & 34.0     & 71.6/94.2 & 79.6/90.6 & 67.1/76.9 & 60.2        & 41.1      \\
Random Init    & 18.9 (0.5)     & 38.6/83.0 (2.1/1.2) & 25.8/71.5 (0.7/0.5) & 16.0/24.2	(2.1/2.0) & 57.6 (0.3)        & 39.0 (0.5)      \\
\midrule
\multicolumn{7}{c}{Section~\ref{section3:synthetic_tasks}: Previously Proposed Synthetic Tasks}                                  \\ \midrule
LIME            & \textbf{33.2 (0.7)}     & \textbf{73.7/94.0 (0.8/0.1)} & \textbf{76.2/89.6 (1.2/0.6)} & \textbf{49.6/61.1	(1.6/1.3)} & \textbf{58.3 (1.1)}       & \textbf{40.7 (0.4)}      \\
Dyck & 27.1     & 65.9/91.9 & 58.5/83.5 & 50.3/62.4 & 58.8        & 40.4      \\
Nons. Summary   & 32.0     & 68.0/92.7 & 65.2/85.2 & 48.4/60.1 & 57.3        & 39.6      \\ \midrule
\multicolumn{7}{c}{Section~\ref{section4:simpler_synthetic_tasks}: Simpler Synthetic Tasks}                                   \\ \midrule
Set             & \textbf{32.8 (0.5)}     & \textbf{71.7/93.7 (0.9/0.2)} & \textbf{73.6/88.6 (1.0/0.3)} & \textbf{48.2/60.0 (0.5/0.7)} & \textbf{58.8 (0.8)}        & \textbf{40.5 (0.9)}      \\
Identity            & 30.2     & 68.6/93.0 & 69.8/86.8 & 26.1/35.4 & 57.8        & 40.5     \\   \midrule
\end{tabular}
\end{adjustbox}
\end{table}

\newpage

\begin{table}[h]
\caption{Section~\ref{section5:per_param_mean_sd} Per Param Mean/SD initialization results for all downstream tasks (rather than just CNNDM-10K and MTOP which were the only two shown in Table~\ref{section5:table_main}). Baseline results and normal fune-tune results are from Appendix Table~\ref{appendix:table_main_with_std}. For CNNDM-10K, MTOP, and SQuAD we provide the mean/SD for five seeds, where for each seed we ran the initialization with a different pre-trained model with different generated data. Per Param Mean/SD initialization gave CNNDM-10K, MTOP, SQuAD, and WebQSP noticeable gains over Random Init, close to gains from Full Init synthetic pre-trained models.} 
\label{appendix:table_per_param_mean_sd}

\begin{adjustbox}{width=1.0\linewidth}
\begin{tabular}{@{}cccccccc@{}}
\midrule
\textbf{}         &           & CNNDM-10K   &           &  &           & MTOP      &           \\
\textbf{}         & LIME      & Set         & Identity  &  & LIME      & Set       & Identity  \\ \midrule
Full Init  & 33.2 (0.7)      & 32.8 (0.5)        & 30.2      &  & 73.7/94.0 (0.8/0.1) & 71.7/93.7 (0.9/0.2) & 68.6/93.0 \\
Per Param Mean/SD & 28.7 (0.7)      & 29.4 (0.6)        & 29.2 (0.6) &  & 67.9/92.6 (3.3/0.8) & 67.3/92.7 (2.5/0.6)  & 66.6/92.2 (1.4/0.5) \\
                  &           &             &           &  &           &           &           \\ \midrule
Baseline          &           & CNNDM-10K   &           &  &           & MTOP      &           \\ \midrule
Random Init       &           & 18.9 (0.5)       &           &  &           & 38.6/83.0 (2.1/1.2) &           \\
Pre-trained T5    &           & 35.8 (0.1)        &           &  &           & 81.0/95.2 (0.3/2.2) &           \\ 
                  &           &             &           &  &           &           &           \\
                  &           &             &           &  &           &           &           \\ \midrule
                  &           & SQuAD       &           &  &           & WebQSP    &           \\
                  & LIME      & Set         & Identity  &  & LIME      & Set       & Identity  \\ \midrule
Full Init  & 49.6/61.1 (1.6/1.3) & 48.2/60.0 (0.5/0.7)   & 26.1/35.4 &  & 76.2/89.6 (1.2/0.6) & 73.6/88.6 (1.0/0.3) & 69.8/86.8 \\
Per Param Mean/SD & 35.4/45.5 (9.2/10.4) & 31.1/40.4 (6.5/7.4)  & 25.6/34.4 (4.9/5.7) &  & 60.9/84.8 & 63.0/86.1 & 58.9/83.6 \\ 
                  &           &             &           &  &           &           &           \\ \midrule
Baseline          &           & SQuAD       &           &  &           & WebQSP    &           \\ \midrule
Random Init       &           & 16.0/24.2 (2.1/2.0)   &           &  &           & 25.8/71.5 (0.7/0.5) &           \\
Pre-trained T5    &           & 77.4/86.1 (0.2/0.2)  &           &  &           & 82.3/91.7 (0.8/0.4) &           \\ 
                  &           &             &           &  &           &           &           \\
                  &           &             &           &  &           &           &           \\ \midrule
                  &           & Code Trans. &           &  &           & Retrosyn. &           \\ \midrule
                  & LIME      & Set         & Identity  &  & LIME      & Set       & Identity  \\
Full Init  & 58.3 (1.1)      & 58.8 (0.8)        & 57.8      &  & 40.7 (0.4)      & 40.5 (0.9)      & 40.5      \\
Per Param Mean/SD & 58.0      & 58.4        & 57.2      &  & 40.3      & 38.7      & 39.3      \\ 
                  &           &             &           &  &           &           &           \\ \midrule
Baseline          &           & Code Trans. &           &  &           & Retrosyn. &           \\ \midrule
Random Init       &           & 57.6 (0.3)        &           &  &           & 39.0 (0.5)      &           \\
Pre-trained T5    &           & 61.2 (0.7)        &           &  &           & 43.5 (0.3)      &           \\ 
\end{tabular}
\end{adjustbox}
\end{table}


\begin{table}[h]
\caption{Pre-attention layer norm initialization sweep results for all downstream tasks (rather than just CNNDM-10K and MTOP which were the only two shown in Table~\ref{section5:table_pre_attn_ln_sweep}). Baseline results are from Appendix Table~\ref{appendix:table_main_with_std}. Lower pre-attention layer norm value gave noticeable benefits over Random Init for CNNDM-10K, MTOP, WebQSP, and SQuAD.} 
\label{appendix:table_pre_attn_ln_sweep}

\begin{adjustbox}{width=1.0\linewidth}
\begin{tabular}{@{}ccccccc@{}}
\toprule
Pre-attn LN Init Value & CNNDM-10K & MTOP      & WebQSP    & SQuAD     & Code Trans. & Retrosyn. \\ \midrule
0.05                   & 28.2      & 25.7/73.3 & 30.5/71.2 & 22.8/31.2      & 57.9        & 39.3      \\
0.1                    & 28.3      & 32.5/77.5 & 28.9/70.6 & 23.5/31.7      & 57.6        & 39.9      \\
0.2                    & 28.2      & 56.4/89.3 & 30.6/72.3 & 28.2/37.3      & 57.0        & 39.4      \\
0.4                    & 24.4      & 56.9/89.9 & 30.0/72.8 & 36.3/46.9      & 57.3        & 39.6      \\
0.8                    & 18.6      & 50.7/88.5 & 28.0/71.6 & 34.2/44.6      & 57.2        & 39.9      \\
                       &           &           &           &           &             &           \\ \midrule
Baseline               & CNNDM-10K & MTOP      & WebQSP    & SQuAD     & Code Trans. & Retrosyn. \\ \midrule
Random Init            & 18.9 (0.5)      & 38.6/83.0 (2.1/1.2) & 25.8/71.5 (0.7/0.5)      & 16.0/24.2 (2.1/2.0) & 57.6 (0.3)        & 39.0 (0.5)      \\
Pre-trained T5         & 35.8 (0.1)      & 81.0/95.2 (0.3/2.2) & 82.3/91.7 (0.8/0.4)      & 77.4/86.1 (0.2/0.2) & 61.2 (0.7)        & 43.5 (0.3)      \\ \bottomrule
\end{tabular}
\end{adjustbox}

\end{table}

\newpage
\section{T5 Small Parameters List}
\label{appendix:t5_small_parameters_list}
\begin{lstlisting}[breaklines]
decoder/decoder_norm/scale
decoder/layers_0/encoder_decoder_attention/key/kernel
decoder/layers_0/encoder_decoder_attention/out/kernel
decoder/layers_0/encoder_decoder_attention/query/kernel
decoder/layers_0/encoder_decoder_attention/value/kernel
decoder/layers_0/mlp/wi/kernel
decoder/layers_0/mlp/wo/kernel
decoder/layers_0/pre_cross_attention_layer_norm/scale
decoder/layers_0/pre_mlp_layer_norm/scale
decoder/layers_0/pre_self_attention_layer_norm/scale
decoder/layers_0/self_attention/key/kernel
decoder/layers_0/self_attention/out/kernel
decoder/layers_0/self_attention/query/kernel
decoder/layers_0/self_attention/value/kernel
decoder/layers_1/encoder_decoder_attention/key/kernel
decoder/layers_1/encoder_decoder_attention/out/kernel
decoder/layers_1/encoder_decoder_attention/query/kernel
decoder/layers_1/encoder_decoder_attention/value/kernel
decoder/layers_1/mlp/wi/kernel
decoder/layers_1/mlp/wo/kernel
decoder/layers_1/pre_cross_attention_layer_norm/scale
decoder/layers_1/pre_mlp_layer_norm/scale
decoder/layers_1/pre_self_attention_layer_norm/scale
decoder/layers_1/self_attention/key/kernel
decoder/layers_1/self_attention/out/kernel
decoder/layers_1/self_attention/query/kernel
decoder/layers_1/self_attention/value/kernel
decoder/layers_2/encoder_decoder_attention/key/kernel
decoder/layers_2/encoder_decoder_attention/out/kernel
decoder/layers_2/encoder_decoder_attention/query/kernel
decoder/layers_2/encoder_decoder_attention/value/kernel
decoder/layers_2/mlp/wi/kernel
decoder/layers_2/mlp/wo/kernel
decoder/layers_2/pre_cross_attention_layer_norm/scale
decoder/layers_2/pre_mlp_layer_norm/scale
decoder/layers_2/pre_self_attention_layer_norm/scale
decoder/layers_2/self_attention/key/kernel
decoder/layers_2/self_attention/out/kernel
decoder/layers_2/self_attention/query/kernel
decoder/layers_2/self_attention/value/kernel
decoder/layers_3/encoder_decoder_attention/key/kernel
decoder/layers_3/encoder_decoder_attention/out/kernel
decoder/layers_3/encoder_decoder_attention/query/kernel
decoder/layers_3/encoder_decoder_attention/value/kernel
decoder/layers_3/mlp/wi/kernel
decoder/layers_3/mlp/wo/kernel
decoder/layers_3/pre_cross_attention_layer_norm/scale
decoder/layers_3/pre_mlp_layer_norm/scale
decoder/layers_3/pre_self_attention_layer_norm/scale
decoder/layers_3/self_attention/key/kernel
decoder/layers_3/self_attention/out/kernel
decoder/layers_3/self_attention/query/kernel
decoder/layers_3/self_attention/value/kernel
decoder/layers_4/encoder_decoder_attention/key/kernel
decoder/layers_4/encoder_decoder_attention/out/kernel
decoder/layers_4/encoder_decoder_attention/query/kernel
decoder/layers_4/encoder_decoder_attention/value/kernel
decoder/layers_4/mlp/wi/kernel
decoder/layers_4/mlp/wo/kernel
decoder/layers_4/pre_cross_attention_layer_norm/scale
decoder/layers_4/pre_mlp_layer_norm/scale
decoder/layers_4/pre_self_attention_layer_norm/scale
decoder/layers_4/self_attention/key/kernel
decoder/layers_4/self_attention/out/kernel
decoder/layers_4/self_attention/query/kernel
decoder/layers_4/self_attention/value/kernel
decoder/layers_5/encoder_decoder_attention/key/kernel
decoder/layers_5/encoder_decoder_attention/out/kernel
decoder/layers_5/encoder_decoder_attention/query/kernel
decoder/layers_5/encoder_decoder_attention/value/kernel
decoder/layers_5/mlp/wi/kernel
decoder/layers_5/mlp/wo/kernel
decoder/layers_5/pre_cross_attention_layer_norm/scale
decoder/layers_5/pre_mlp_layer_norm/scale
decoder/layers_5/pre_self_attention_layer_norm/scale
decoder/layers_5/self_attention/key/kernel
decoder/layers_5/self_attention/out/kernel
decoder/layers_5/self_attention/query/kernel
decoder/layers_5/self_attention/value/kernel
decoder/relpos_bias/rel_embedding
encoder/encoder_norm/scale
encoder/layers_0/attention/key/kernel
encoder/layers_0/attention/out/kernel
encoder/layers_0/attention/query/kernel
encoder/layers_0/attention/value/kernel
encoder/layers_0/mlp/wi/kernel
encoder/layers_0/mlp/wo/kernel
encoder/layers_0/pre_attention_layer_norm/scale
encoder/layers_0/pre_mlp_layer_norm/scale
encoder/layers_1/attention/key/kernel
encoder/layers_1/attention/out/kernel
encoder/layers_1/attention/query/kernel
encoder/layers_1/attention/value/kernel
encoder/layers_1/mlp/wi/kernel
encoder/layers_1/mlp/wo/kernel
encoder/layers_1/pre_attention_layer_norm/scale
encoder/layers_1/pre_mlp_layer_norm/scale
encoder/layers_2/attention/key/kernel
encoder/layers_2/attention/out/kernel
encoder/layers_2/attention/query/kernel
encoder/layers_2/attention/value/kernel
encoder/layers_2/mlp/wi/kernel
encoder/layers_2/mlp/wo/kernel
encoder/layers_2/pre_attention_layer_norm/scale
encoder/layers_2/pre_mlp_layer_norm/scale
encoder/layers_3/attention/key/kernel
encoder/layers_3/attention/out/kernel
encoder/layers_3/attention/query/kernel
encoder/layers_3/attention/value/kernel
encoder/layers_3/mlp/wi/kernel
encoder/layers_3/mlp/wo/kernel
encoder/layers_3/pre_attention_layer_norm/scale
encoder/layers_3/pre_mlp_layer_norm/scale
encoder/layers_4/attention/key/kernel
encoder/layers_4/attention/out/kernel
encoder/layers_4/attention/query/kernel
encoder/layers_4/attention/value/kernel
encoder/layers_4/mlp/wi/kernel
encoder/layers_4/mlp/wo/kernel
encoder/layers_4/pre_attention_layer_norm/scale
encoder/layers_4/pre_mlp_layer_norm/scale
encoder/layers_5/attention/key/kernel
encoder/layers_5/attention/out/kernel
encoder/layers_5/attention/query/kernel
encoder/layers_5/attention/value/kernel
encoder/layers_5/mlp/wi/kernel
encoder/layers_5/mlp/wo/kernel
encoder/layers_5/pre_attention_layer_norm/scale
encoder/layers_5/pre_mlp_layer_norm/scale
encoder/relpos_bias/rel_embedding
\end{lstlisting}

\newpage

\begin{figure}[h]
\centering 
\centerline{\includegraphics[width=0.7\columnwidth]{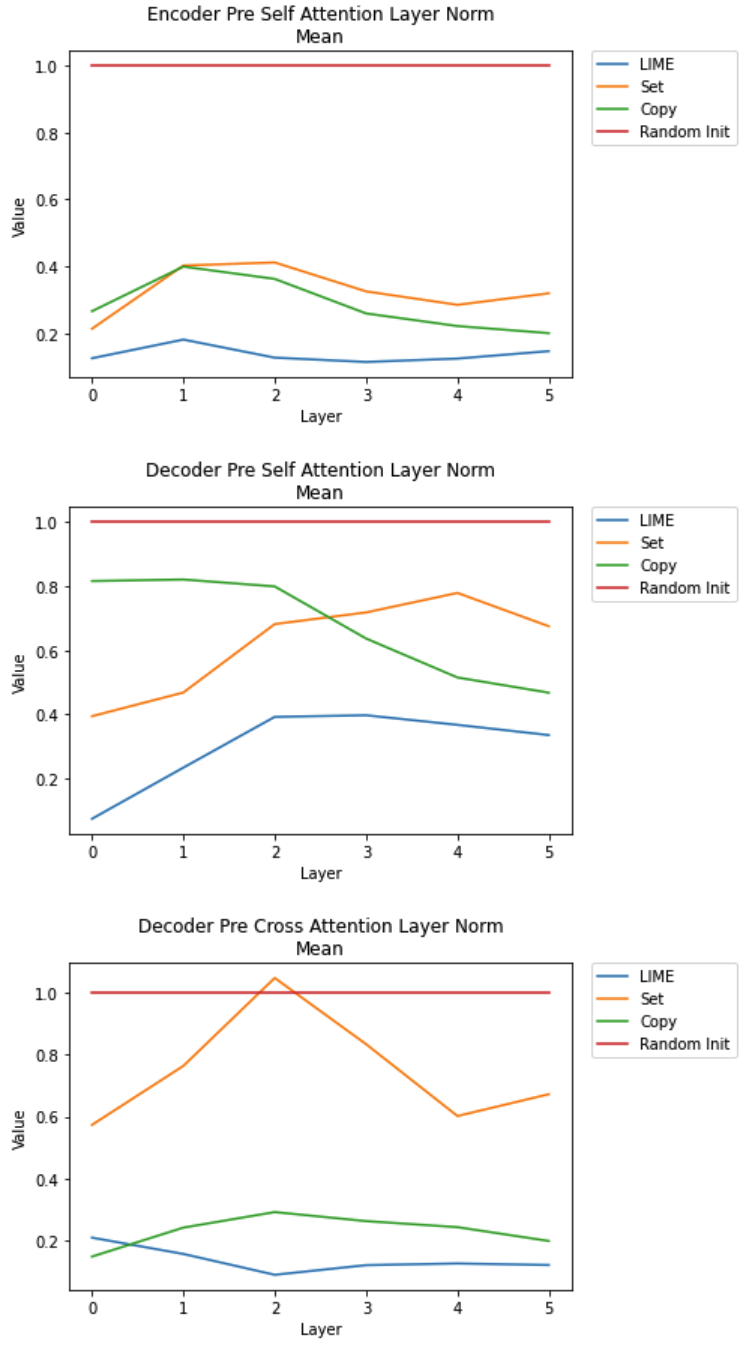}}
\caption{For each layer, compute the mean over weights of the pre-attention layer norm parameter in that layer. Compute and plot these values for a Random Init model as well as LIME, Set, and Identity pre-trained models. For example, looking at the top-most plot, the mean of the Set pre-trained model's encoder layer 0 pre-attention layer norm is about 0.2. And the mean of the Identity pre-trained model's encoder layer 1 pre-attention layer norm is about 0.4. Looking at the bottom-most plot, the mean of the Set pre-trained model's encoder layer 4 pre-cross-attention layer norm is about 0.6.} 
\label{appendix:figure_plot_pre_attn_ln}
\end{figure}
\end{document}